\documentclass[letterpaper, 10 pt, conference, twoside]{ieeeconf}

\IEEEoverridecommandlockouts                              

\ifdefined\pdfminorversion
\fi

\usepackage{graphics} 
\usepackage{epsfig} 
\usepackage{times} 

\usepackage{float}

\let\labelindent\relax
\usepackage{enumitem}
\usepackage{cite}
\usepackage{booktabs}
\usepackage{amsmath,amssymb,amsfonts}
\usepackage{algorithmic}
\usepackage{setspace}
\usepackage{subcaption} 
\usepackage{graphicx}
\usepackage{textcomp}
\usepackage[table,xcdraw]{xcolor}
\usepackage{pdfpages} 
\usepackage{multirow}
\usepackage{threeparttable}
\usepackage{tabularx}
\usepackage{mathtools}
\usepackage{caption}
\usepackage{pifont}
\usepackage{nameref}
\usepackage{wrapfig}
\usepackage{placeins}
\usepackage{colortbl}

\definecolor{gradient1}{RGB}{187, 210, 160}
\definecolor{gradient2}{RGB}{227, 230, 189}
\definecolor{gradient3}{RGB}{253, 240, 207}
\definecolor{gradient4}{RGB}{254, 232, 204}
\definecolor{gradient5}{RGB}{246, 206, 190}
\definecolor{gradient6}{RGB}{231, 178, 172}

\definecolor{g1}{HTML}{1A9850}
\definecolor{g2}{HTML}{66BD63}
\definecolor{g3}{HTML}{A6D96A}
\definecolor{g4}{HTML}{FEE08B}
\definecolor{g5}{HTML}{FEE08B}
\definecolor{g6}{HTML}{FDAE61}
\definecolor{g7}{HTML}{F46D43}
\definecolor{g8}{HTML}{F46D43}
\definecolor{g9}{HTML}{D73027}

\definecolor{mydarkblue}{rgb}{0,0.08,0.45}
\newcommand{\logo}[1]{\raisebox{-0.25ex}{\includegraphics[height=2ex]{#1}}\hspace{3pt}}
\newcommand{\airiob}{AirIO$^{\dagger}$} 
\definecolor{dgreen}{RGB}{0,160,40}   
\definecolor{dgrey}{RGB}{140,140,140}  
\definecolor{oursbg}{RGB}{255,240,219}  
\newcommand{\dlt}[2]{{\scriptsize\textcolor{#2}{(#1)}}} 
\newcommand{\dcell}[3]{\begin{tabular}[t]{@{}c@{}}#1\\[-1pt]\dlt{#2}{#3}\end{tabular}} 

\usepackage[free-standing-units=true]{siunitx}
\newcommand{\rom}[1]{\uppercase\expandafter{\romannumeral #1\relax}}
\usepackage{array}
\newcommand{\PreserveBackslash}[1]{\let\temp=\\#1\let\\=\temp}
\newcolumntype{C}[1]{>{\PreserveBackslash\centering}p{#1}}
\newcolumntype{R}[1]{>{\PreserveBackslash\raggedleft}p{#1}}
\newcolumntype{L}[1]{>{\PreserveBackslash\raggedright}p{#1}}
\newcommand{\fref}[1]{Fig.~\ref{#1}}
\newcommand{\sref}[1]{Sec.~\ref{#1}}
\newcommand{\tref}[1]{Tab.~\ref{#1}}
\newcommand{\eref}[1]{Eq.~\ref{#1}}
\newcommand{\aref}[1]{Appendix~\ref{#1}}

\newcommand{\citep}[1]{\cite{#1}}
\newcommand{\citet}[1]{\cite{#1}}

\makeatletter
\let\NAT@parse\undefined
\makeatother

\usepackage[colorlinks]{hyperref}
  \hypersetup{
    citecolor=blue,
    linkcolor=blue,   
    urlcolor=blue}

\begin{document}

\title{\LARGE \bf Self-Supervised Inertial Odometry\\[2pt]
  {\normalsize\normalfont\href{https://SSIO-IMU.github.io}{SSIO-IMU.github.io}}\vspace{8pt}}

\author{
  \parbox{\dimexpr\textwidth-2\tabcolsep}{\centering\normalfont
  {\bfseries
  Yuheng~Qiu$^{1,2}$, Yutian~Chen$^{1,2}$, Carlton~Cort$^{2}$, Weison~Huang$^{2}$, Shibo~Zhao$^{1}$,\\
  Pieter~Abbeel$^{1,3}$, Rocky~Duan$^{1}$, Koushil~Sreenath$^{1,3}$, C.~Karen~Liu$^{1,4}$,\\ Carmelo~Sferrazza$^{1}$,
  Junyi~Geng$^{5}$, Sebastian~Scherer$^{2}$, Guanya~Shi$^{1,2}$}\\
  $^{1}$Amazon FAR, $^{2}$Carnegie Mellon University, $^{3}$UC Berkeley,
  $^{4}$Stanford University, $^{5}$Penn State University
  }}

\thispagestyle{empty}
\pagestyle{headings}
\makeatletter
\g@addto@macro\@maketitle{
  \captionsetup{type=figure}\setcounter{figure}{0}
  \def\mycolspace{1.2mm}
  \centering
  \includegraphics[width=\linewidth]{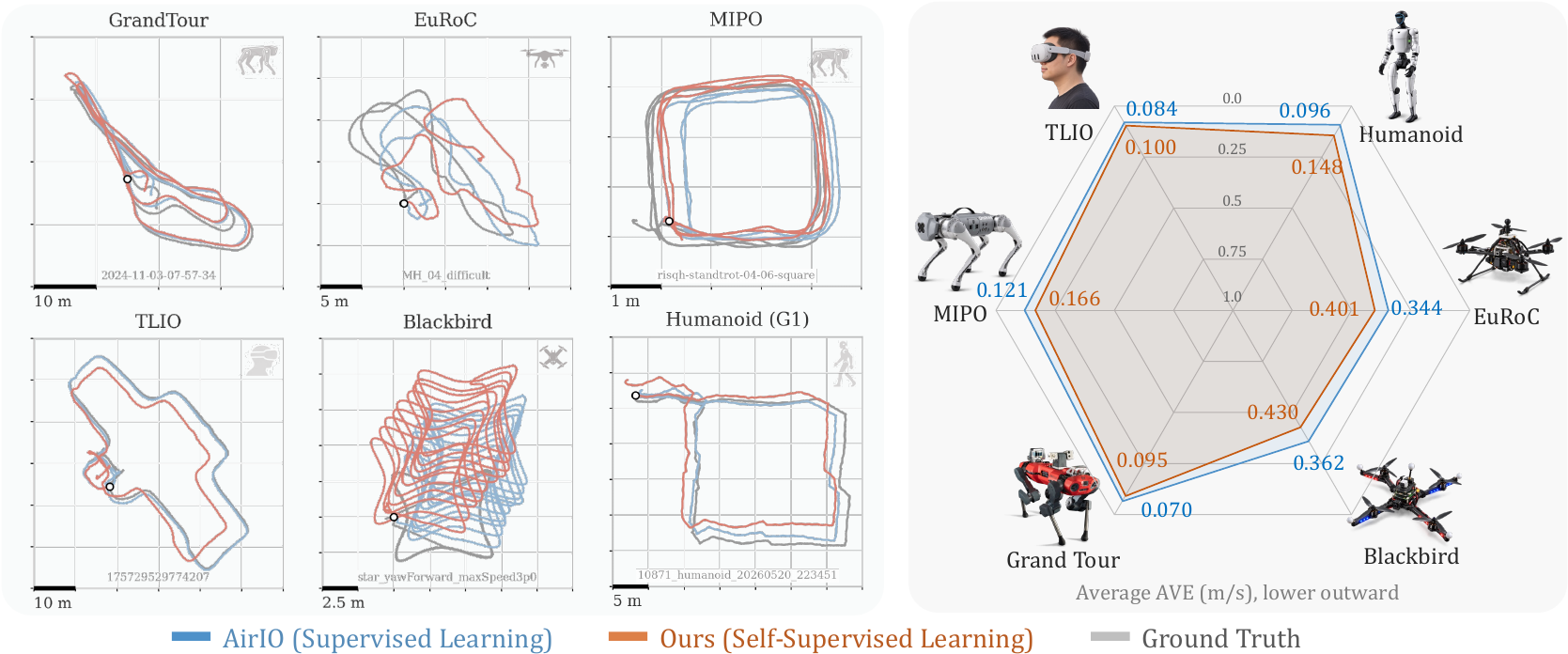}
  \captionof{figure}{
  \textbf{Learning inertial odometry without velocity or position labels across six datasets.}
  Models are trained separately from scratch.
  With a reference attitude for teacher gravity compensation, our models refined by pose graph distillation (PGD, {\color{red}red}) have an average velocity error within $0.07$~m/s of the same-architecture supervised \airiob{} baseline ({\color{blue}blue}).
  }
  \label{fig:overview}
}

\makeatother

\maketitle
\begin{abstract}
Learning inertial odometry for a new platform usually requires ground-truth trajectories.
We show that an inertial odometry model can instead be trained from scratch by a \textit{myopic} teacher that integrates the IMU over each short window independently, a scheme we call \textit{integration self-supervision}.
Interestingly, although supervised only by short-window velocity changes, the student learns to directly predict velocity, and does so more accurately than its integration teacher.
Pose graph distillation then refines the model with sequence-consistent targets from a pose graph over many windows.
Training on a new platform therefore no longer requires collecting ground-truth trajectories, but only calibrating the IMU and recording data during normal operation.
Across six datasets spanning pedestrians, drones, quadrupeds, and a humanoid, the same recipe with no velocity or position labels trains models that come within $0.07$~m/s average velocity error of a same-architecture supervised baseline.
\end{abstract}

\section{Introduction}
\label{sec:introduction}

Inertial odometry (IO) is self-contained, inexpensive, and remains available when cameras, LiDAR, or GNSS become unreliable or unavailable.
Its main limitation is drift, as double integration accumulates IMU noise and bias into rapidly growing position error.
To reduce this drift, existing methods exploit motion structure obtained from analytical models or trained from labeled data.
Model-based approaches derive motion constraints from platform kinematics or dynamics, such as leg kinematics~\citep{yang2023mipo, wu2025doglegs}.
Learning-based IO instead learns motion priors from pose- or velocity-labeled trajectories~\citep{qiu2025airio, chen2018ionet, yan2018ridi, herath2020ronin, liu2020tlio}.
Whether explicitly through kinematics or implicitly through labeled data, both approaches encode embodiment, so the resulting model is tied to one geometry, sensor mounting, payload, and actuation.
Adapting to a new embodiment thus requires a new analytical model or new labeled trajectories, limiting cross-embodiment generalization.

For learning-based IO, this generalization failure creates a labeling bottleneck.
Ground-truth velocity or pose requires motion capture, RTK, or SLAM, which are costly to deploy and often degrade or become unavailable in complex real-world scenarios.
The burden compounds at deployment because every new embodiment, payload, mounting, or control policy forces re-collecting ground-truth trajectories on the new platform~\citep{qiu2025airio, zhao2025tartan}.
Meanwhile, robots generate abundant unlabeled IMU data during operation that supervised approaches cannot exploit.
Lifting the label requirement would turn every deployed robot into its own source of training data, enable embodiments where ground truth is infeasible, and let the model continue learning as mounting, payload, and policy drift from training conditions.
This raises the question:
\par\nopagebreak\vspace{2pt}
{\centering\emph{\textbf{\quad Can a robot learn its own motion model from IMU data only?}}\par}
\vspace{2pt}
Our thesis is that a \textit{myopic integration teacher} can train an inertial odometry student that learns the embodiment's motion model from unlabeled IMU data and becomes more accurate than the teacher.
The teacher integrates accelerometer and gyroscope measurements into rotation, velocity, and position increments over short windows~\citep{forster2015imu}.
It is myopic because it integrates each window separately, so it carries no memory across windows and no embodiment-specific prior beyond the IMU integration equations, and its increments are noisy.
The student, in contrast, sees a longer IMU context and is trained on thousands of such windows.
Teacher errors vary across windows while the embodiment's motion repeats, so the student learns the repeated motion patterns, such as gait periodicity or thrust cycles, and averages the noise away.

Grounded in this insight, we introduce a self-supervised method that learns inertial odometry from unlabeled IMU data through \textit{integration self-supervision}.
Integrating the raw IMU over a short window gives the body-frame velocity change between its endpoints.
The network is trained so that its predicted endpoint velocities differ by this change.
Once this signal yields reliable velocity estimates, \textit{pose graph distillation} (PGD) can further refine the network: one pose graph links the integrated velocity changes of many consecutive windows with the network's own predictions, and its solution provides sequence-consistent velocity targets for fine-tuning.
Generalizing to a new embodiment therefore requires no new labels and no new analytical model.
Instead of collecting ground truth with motion capture, RTK, or instrumented rigs, we only calibrate the IMU and record data during normal operation.
Extra proprioceptive streams such as additional IMUs or joint encoders extend the same losses to a multi-modal student and strengthen the student.
Across six datasets, self-supervised training approaches the supervised baseline, and on TLIO its accuracy improves monotonically as more unlabeled data become available.
In summary, our contributions are:
\begin{itemize}[leftmargin=*,itemsep=2pt,topsep=2pt]
    \item We introduce a self-supervised framework for inertial odometry that learns an embodiment-specific motion model from unlabeled IMU data.
    \item To the best of our knowledge, we are the first to show that training from scratch on unlabeled IMU data alone can approach the performance of a supervised learning-based IO model.
    \item We show that the student exceeds its integration teacher over long horizons, specializes to its embodiment, and improves as unlabeled training data scale up.
\end{itemize}

\section{Related Work}
\label{sec:related}
\begin{figure}[t]
\centering
\includegraphics[width=\linewidth]{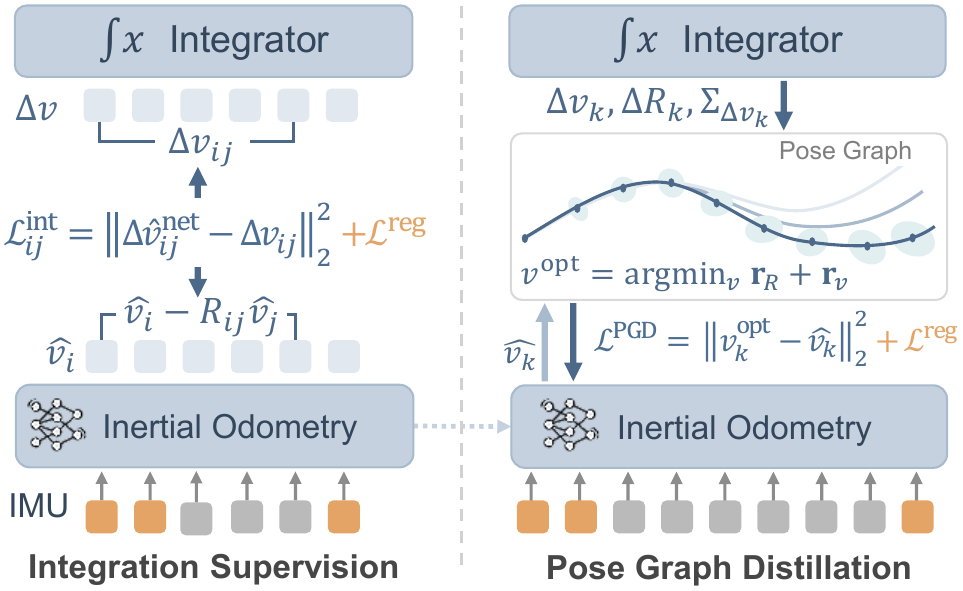}
\caption{\textbf{Integration Self-Supervision} trains the motion network to match the body-frame velocity change predicted by the frozen integration teacher under a squared velocity-change loss, supervising only the velocity mean. \textbf{Pose Graph Distillation (PGD)} builds a full-sequence pose graph from the teacher's IMU factors and the network's velocity factors. The graph posterior $\mathbf{v}^{\mathrm{opt}}$ then supervises the network through an L2 loss. \textbf{Velocity Regularization}, $\mathcal{L}^{\mathrm{reg}}$, is used over detected static frames.} 
\label{fig:SSL_training}
\end{figure}

\subsection{Learning-Based Inertial and Legged Odometry}
\label{sec:rw_learned_io}

Early neural IO systems showed that IMU windows contain motion cues beyond what classical dead reckoning can recover.
IONet~\citep{chen2018ionet}, RIDI~\citep{yan2018ridi}, RoNIN~\citep{herath2020ronin}, and TLIO~\citep{liu2020tlio} learn velocity from raw IMU sequences for pedestrians, and later systems extend learned IO to drones and other robots through new representations or dynamics models~\citep{cioffi2023learned, zhang2022dido, qiu2025airio}.
Many of these systems also expose uncertainty for downstream filters and graphs, through classical IMU integration~\citep{forster2015imu}, learned velocity covariance~\citep{liu2020tlio}, or learned bias and correction terms~\citep{buchanan2022deep, qiu2023airimu}.
A complementary line keeps a model-based estimator and learns only its corrections, such as noise adaptation inside a Kalman filter~\citep{brossard2020aiimu} or gyroscope denoising~\citep{brossard2020denoising}, while geometric priors such as rotation equivariance~\citep{cao2022rio, jayanth2025eqnio} improve how a learned model generalizes across orientations.
On articulated platforms, classical multi-IMU filters such as MIPO~\citep{yang2023mipo} and DogLegs~\citep{wu2025doglegs} fuse body IMU, joint encoders, and additional leg-mounted IMUs to detect contacts and constrain leg kinematics in an EKF, while learning-based proprioceptive odometry systems such as Buchanan et al.~\citep{buchanan2022learning}, Legolas~\citep{wasserman2024legolas}, and AutoOdom~\citep{luo2025autoodom} regress velocity from the same multi-modal inputs.
All of these learning-based systems rely on supervised velocity, pose, or simulation labels to fit the embodiment-specific motion prior, whereas our method learns from unlabeled IMU data on the target embodiment and reuses the same recipe across body-IMU and proprioceptive inputs.

\subsection{Self-Supervision in Inertial Navigation}
\label{sec:rw_self_supervised}

Prior methods self-supervise the odometry model in two ways.
The first borrows a second sensor.
DeepVIO~\citep{han2019deepvio} trains its IMU branch from stereo geometric constraints, KISS-IMU~\citep{choi2026kissimu} supervises the network with LiDAR ICP and pose-graph optimization, and self-supervised visual ego-motion learns from photometric consistency~\citep{zhou2017unsupervised}.
iSLAM~\citep{fu2024islam} and imperative learning~\citep{wang2024imperative} generalize this to a bilevel structure in which an optimizer over the sensor's residuals teaches a learned front end through differentiable optimization.
All of these trade the label rig for a camera or LiDAR, so they still require a sensor the target platform may not carry.
The second leverages a prior on the model.
RIO~\citep{cao2022rio} uses rotation equivariance as an auxiliary signal and EqNIO~\citep{jayanth2025eqnio} builds equivariance into the architecture, but both still fit the motion prior from labels and use the invariance only to reduce how many are needed.
Imperative MPC~\citep{he2025imperative} instead self-supervises the network through a differentiable model predictive controller, so its supervision rests on a known dynamics model of the platform.
Our method needs neither an additional sensor nor such a prior, because the IMU integrator itself supplies the training target.

\section{Self-Supervised Inertial Odometry}
\label{sec:method}

We train an inertial odometry model for a target platform using only its unlabeled onboard measurements (\fref{fig:SSL_training}).
The foundation of our method is integration self-supervision: the IMU measurements inside each sampled window are integrated into a body-frame velocity change between its endpoints, and the network is trained so that the difference of its predicted velocities at those endpoints matches this integrated change.
A regularization loss on IMU-detected static frames suppresses the drift that integration noise and bias induce where the measurements carry no excitation.
Once the network provides reliable velocity estimates, PGD can refine it with velocities optimized jointly over many windows in a pose graph.
Every network is trained from random initialization on unlabeled training sequences, without ground-truth velocity or position as training targets (the initial state of each PGD graph is described in \sref{sec:experiments}), and evaluated on a held-out test split.

\begin{figure}[t]
\centering
\includegraphics[width=0.88\linewidth]{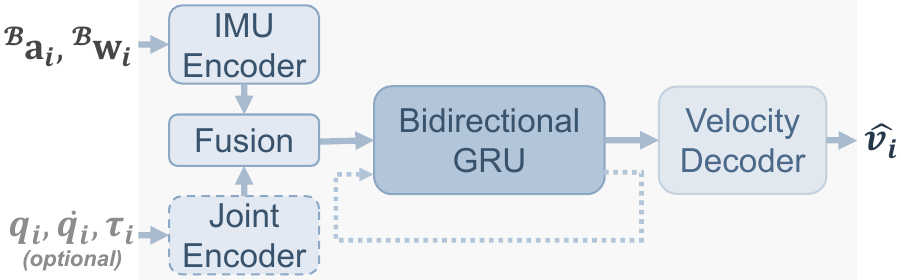}
\caption{\textbf{Motion network} architecture used for both stages of self-supervised training and the supervised \airiob{} baseline.}
\label{fig:motion_network}
\end{figure}

\textbf{Motion network.}
At deployment the student sees only the IMU, with no label and no guaranteed onboard attitude filter, so it must infer velocity from the raw measurements alone.
We therefore build on AirIO~\citep{qiu2025airio} and remove its attitude encoder, denoting the modified network \airiob{} (\fref{fig:motion_network}).
Given the raw accelerometer and gyroscope measurements in a window of $W$ samples ending at sample $k$, the network predicts the body-frame velocity
\begin{equation}
{}^\mathcal{B}\hat{\mathbf{v}}_k=f_{\theta}\big(\{{}^\mathcal{B}\mathbf{a}_t,\,{}^\mathcal{B}\boldsymbol{\omega}_t\}_{t=k-W+1}^{k}\big),
\label{eq:motion_network}
\end{equation}
where $\theta$ are the network weights and the superscript $\mathcal{B}$ denotes the IMU body frame at the corresponding timestamp.
Following the observability insight of AirIO, we keep both the input and the predicted velocity in the body frame.
Raw body-frame measurements retain the gravity direction, which a global-frame, gravity-removed input discards, so the network can infer tilt from the IMU history itself.
The body frame also suits self-supervision, because the teacher's preintegrated velocity change is expressed in the body frame at the window start and can be compared with body-frame predictions through the relative rotation alone, without any global heading.
Removing the attitude encoder costs the supervised \airiob{} less than $0.05$~m/s in average velocity error (AVE) relative to AirIO (\tref{tab:setup2}).

\subsection{Integration Self-Supervision}
\label{sec:m1}
\addtocounter{table}{1}
\begin{table*}[!t]
\centering
\captionsetup{font=small}
\caption{
\textbf{Self-Supervised Learning from Scratch.}
We report AVE (m/s) and ATE (m) across six datasets.
Supervised rows are trained with ground-truth labels and serve as reference; ours are trained from random initialization without velocity or position labels.
}
\label{tab:setup2}
\resizebox{\textwidth}{!}{%
\begin{tabular}{l cc cc cc cc cc cc}
\toprule
 & \multicolumn{2}{c}{\logo{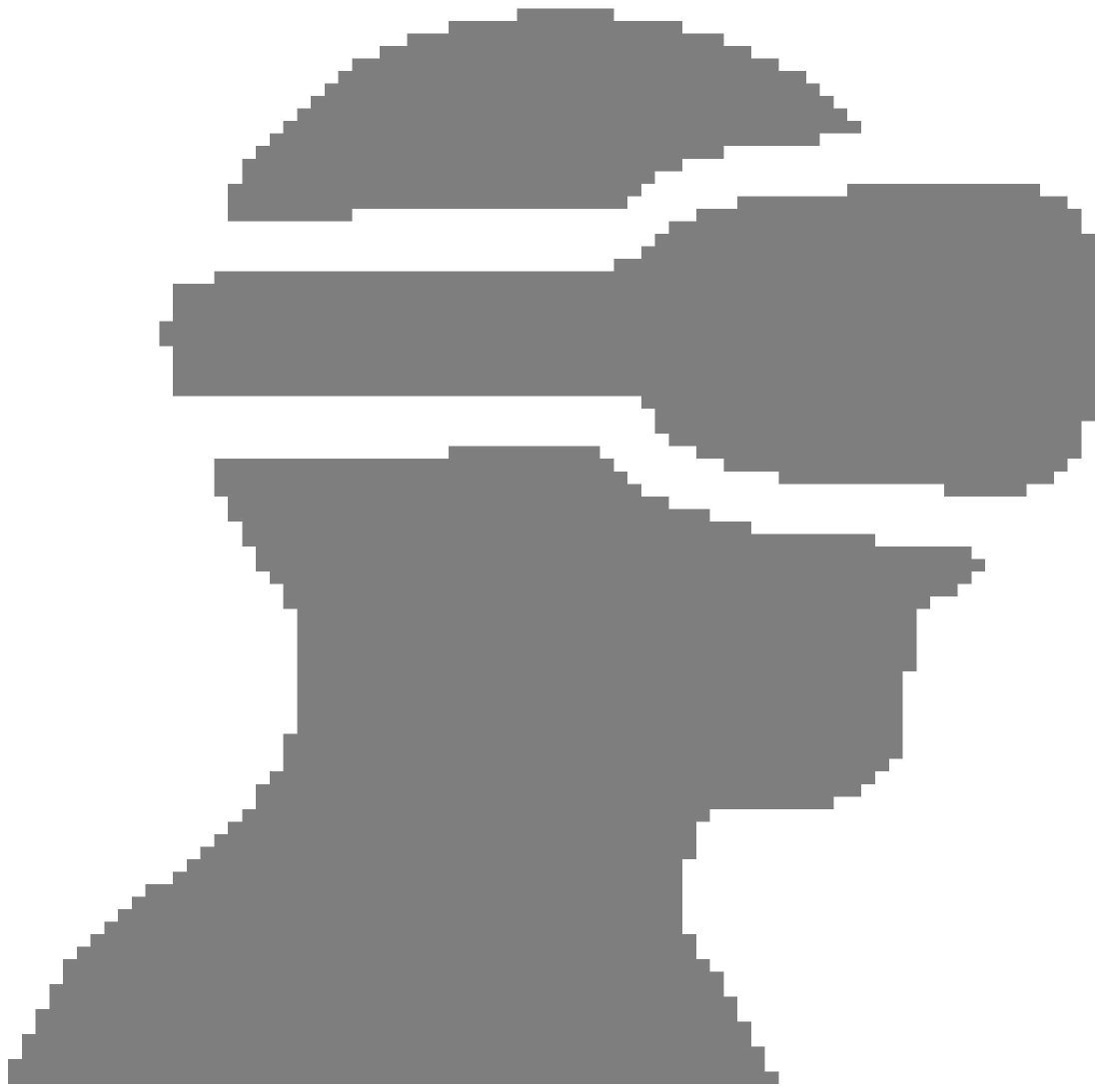}TLIO} & \multicolumn{2}{c}{\logo{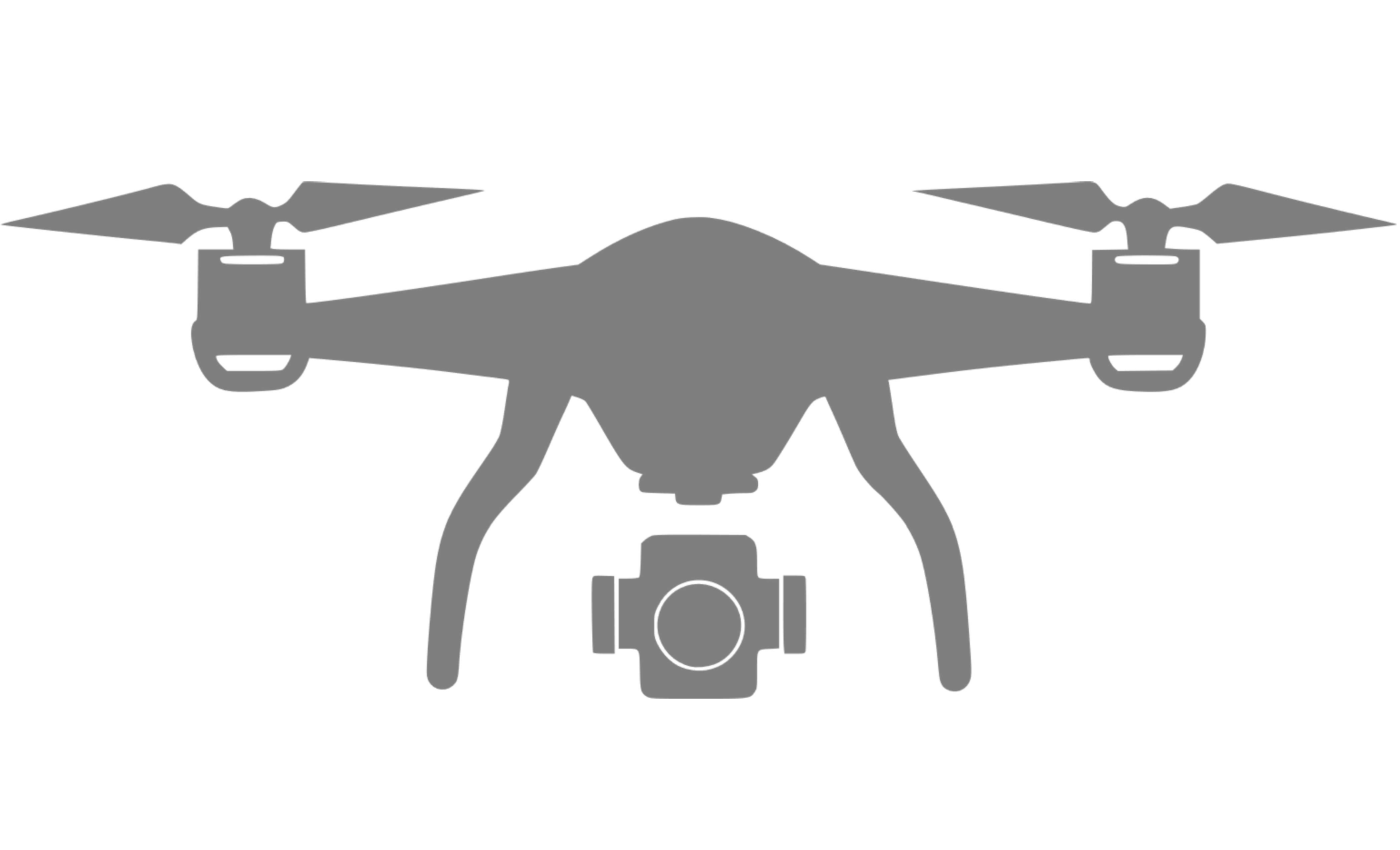}EuRoC} &
 \multicolumn{2}{c}{\logo{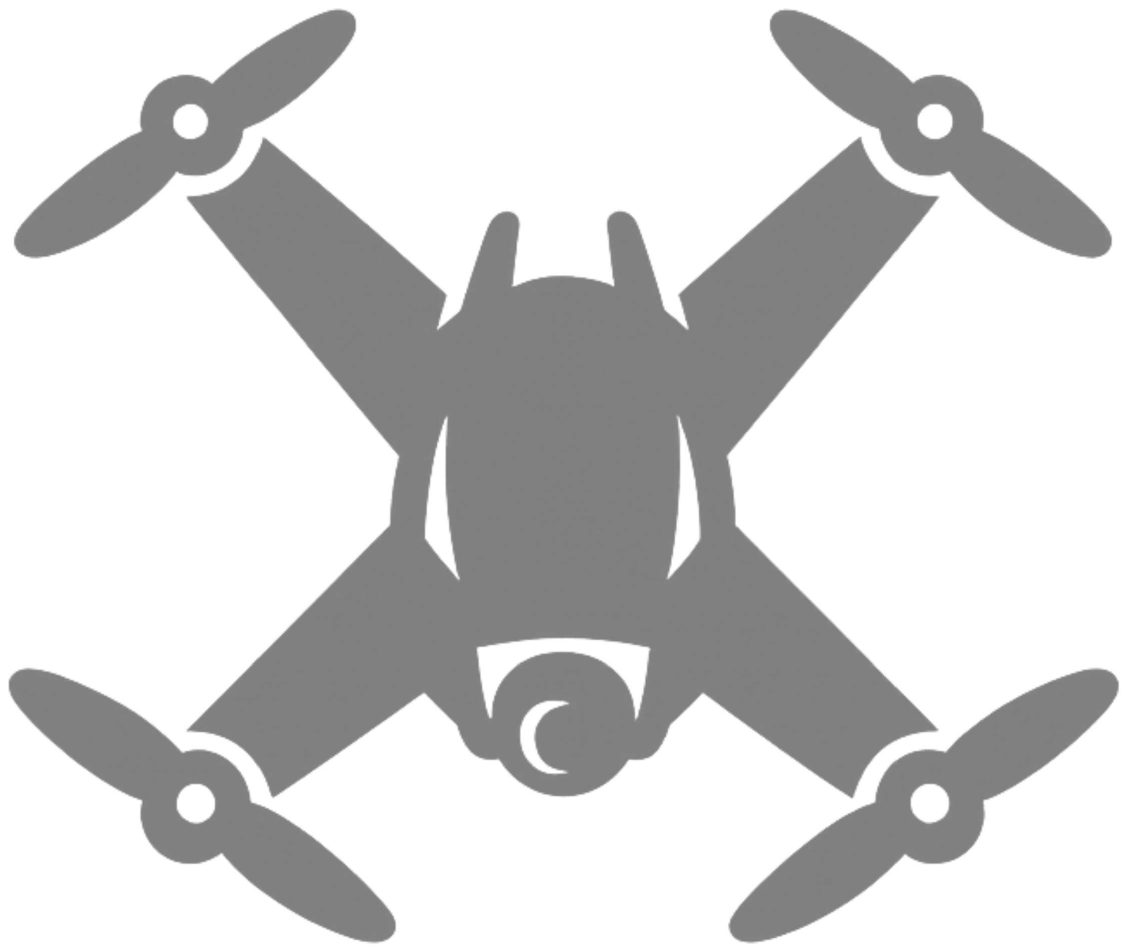}Blackbird}
 & \multicolumn{2}{c}{\logo{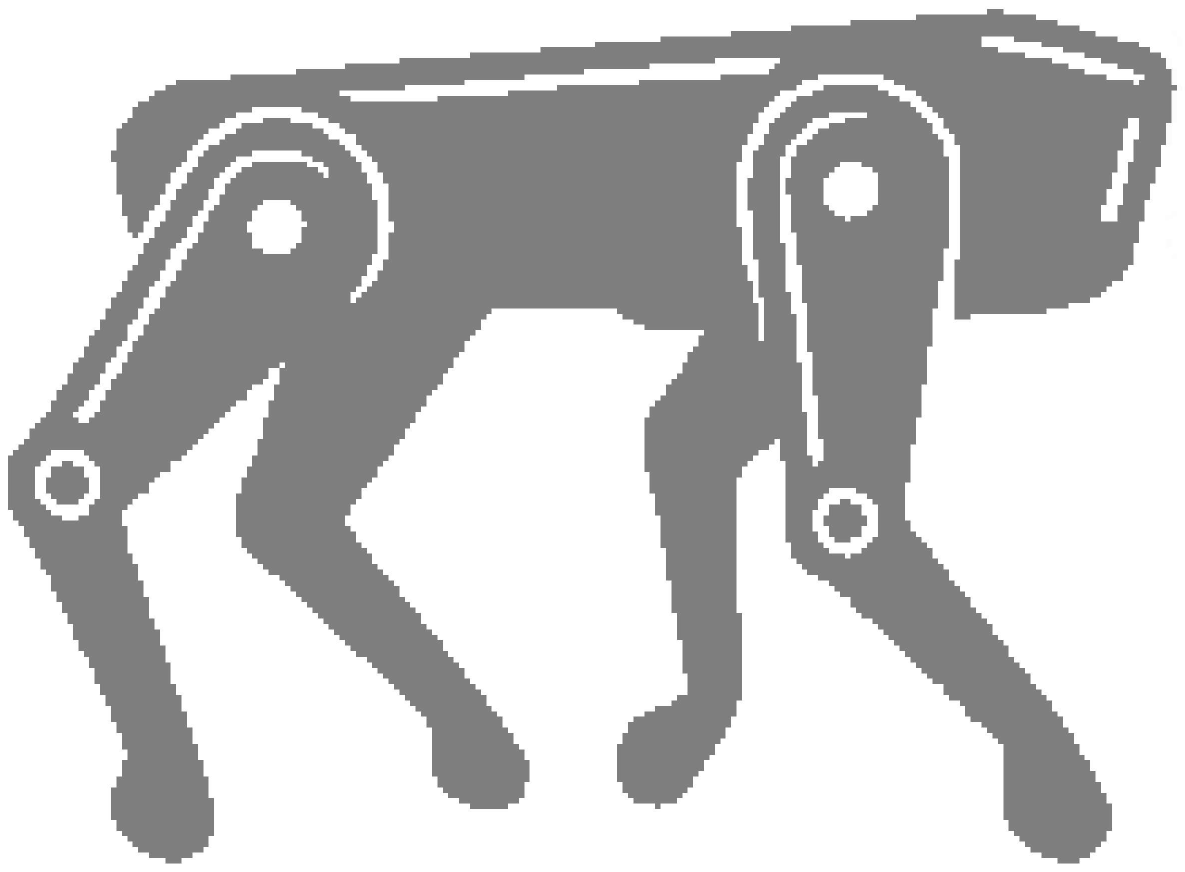}MIPO}
 & \multicolumn{2}{c}{\logo{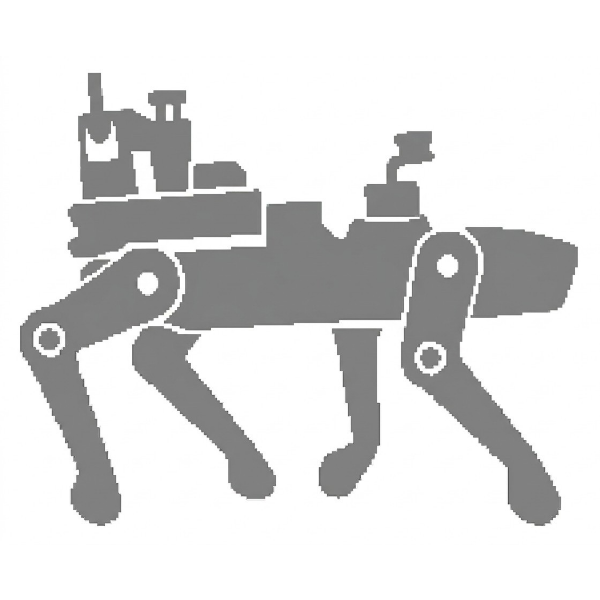}GrandTour}
 & \multicolumn{2}{c}{\logo{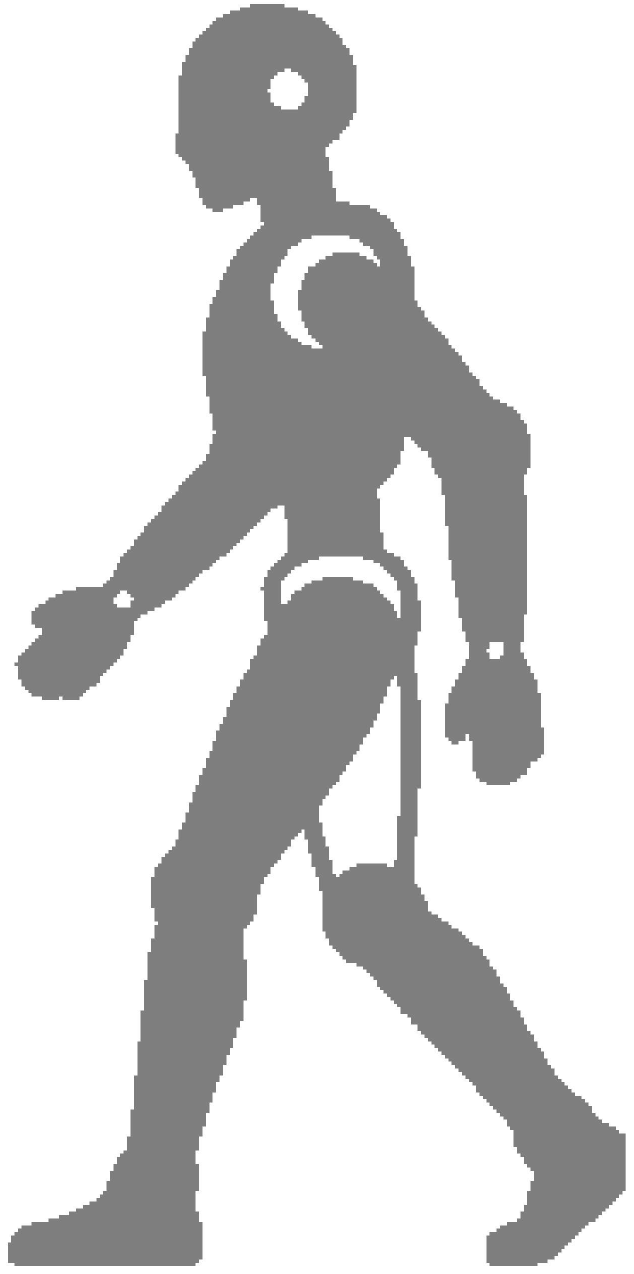}Humanoid} \\
\cmidrule(lr){2-3} \cmidrule(lr){4-5} \cmidrule(lr){6-7} \cmidrule(lr){8-9} \cmidrule(lr){10-11} \cmidrule(lr){12-13}
Method & AVE~$\downarrow$ & ATE~$\downarrow$ & AVE~$\downarrow$ & ATE~$\downarrow$ & AVE~$\downarrow$ & ATE~$\downarrow$ & AVE~$\downarrow$ & ATE~$\downarrow$ & AVE~$\downarrow$ & ATE~$\downarrow$ & AVE~$\downarrow$ & ATE~$\downarrow$ \\
\midrule
\multicolumn{13}{l}{\textbf{Supervised Learning}} \\
\hspace{3mm} AirIMU~\citep{qiu2023airimu}
  & 20.83 & --
  & 13.42 & --
  & 46.95 & --
  & 1.958  & 39.07
  & 2.448  & 334.8
  & 0.618  & -- \\
\hspace{3mm} RoNIN~\citep{herath2020ronin}
  & 0.160  & 2.420
  & 0.661  & 9.011
  & 0.948  & 16.49
  & 0.157  & 0.429
  & 0.102  & 3.894
  & 0.163  & 1.989 \\
\hspace{3mm} TLIO~\citep{liu2020tlio}
  & 0.161  & 2.872
  & 0.656  & 8.541
  & 0.972  & 15.54
  & 0.160  & 0.487
  & 0.106  & 4.327
  & 0.163  & 1.737 \\
\hspace{3mm} AirIO~\citep{qiu2025airio}
  & 0.081  & 1.890
  & 0.335  & 3.720
  & 0.318  & 6.100
  & 0.120  & 0.310
  & 0.068  & 3.190
  & 0.087  & 1.260 \\
\hspace{3mm} \airiob{} - \textit{baseline}
  & 0.084  & 2.080
  & 0.344  & 3.720
  & 0.362  & 6.390
  & 0.121  & 0.310
  & 0.070  & 3.380
  & 0.096  & 1.400 \\
\midrule
\multicolumn{13}{l}{\textbf{Self-Supervised From Scratch (ours)}} \\
\rowcolor{oursbg} \hspace{3mm} Ours
  & \dcell{0.104}{+0.020}{dgreen} & \dcell{3.040}{+0.960}{dgreen}
  & \dcell{0.423}{+0.079}{dgreen} & \dcell{9.030}{+5.310}{dgrey}
  & \dcell{0.475}{+0.113}{dgrey} & \dcell{18.39}{+12.00}{dgrey}
  & \dcell{0.182}{+0.061}{dgreen} & \dcell{1.630}{+1.320}{dgrey}
  & \dcell{0.096}{+0.026}{dgreen} & \dcell{4.370}{+0.990}{dgreen}
  & \dcell{0.167}{+0.071}{dgreen} & \dcell{4.360}{+2.960}{dgrey} \\

\rowcolor{oursbg} \hspace{3mm} Ours (w/ PGD)
  & \dcell{0.100}{+0.016}{dgreen} & \dcell{3.010}{+0.930}{dgreen}
  & \dcell{0.401}{+0.057}{dgreen} & \dcell{10.23}{+6.51}{dgrey}
  & \dcell{0.430}{+0.068}{dgreen} & \dcell{17.93}{+11.54}{dgrey}
  & \dcell{0.166}{+0.045}{dgreen} & \dcell{1.640}{+1.330}{dgrey}
  & \dcell{0.095}{+0.025}{dgreen} & \dcell{4.340}{+0.960}{dgreen}
  & \dcell{0.148}{+0.052}{dgreen} & \dcell{4.270}{+2.870}{dgrey} \\

\bottomrule
\end{tabular}}
\par\vspace{2pt}
{\footnotesize\parbox{\textwidth}{\raggedright
``--'' denotes error too large to report.
Ours denotes integration self-supervision and Ours (w/ PGD) adds the pose graph distillation refinement.
\airiob{} is our customized AirIO model without the attitude encoder.
Parenthesized numbers give the difference to \airiob{}; \textcolor{dgreen}{green} marks a difference below $+0.1$~m/s in AVE or $+1$~m in ATE; \textcolor{dgrey}{gray} a larger one.}}
\vspace{0pt}
\end{table*}

The teacher uses standard IMU preintegration~\citep{forster2015imu} over each short window $(i,j)$ to obtain a relative rotation $\Delta\mathbf{R}_{ij}$ and a gravity-compensated velocity change $\Delta\mathbf{v}_{ij}$, both expressed in the body frame at $i$.
Before integration, we subtract constant accelerometer and gyroscope biases estimated from static frames in each sequence.
A constant bias is the simplest correction that static frames can estimate without labels, and \fref{fig:noise_bias_surface} shows that bias, not noise, is the error the teacher cannot average away.
The teacher requires an estimate of gravity direction and relative rotation; the attitude source used in our experiments is stated in \sref{sec:experiments}.

For each sampled window $(i,j)$, the network outputs the body-frame velocities ${}^\mathcal{B}\hat{\mathbf{v}}_i$ and ${}^\mathcal{B}\hat{\mathbf{v}}_j$ at the two endpoints, from which we form the network's velocity increment in the body frame at $i$,
\begin{equation}
\label{eq:net_increment}
\Delta\hat{\mathbf{v}}^{\mathrm{net}}_{ij}=\Delta\mathbf{R}_{ij}\,{}^\mathcal{B}\hat{\mathbf{v}}_j-{}^\mathcal{B}\hat{\mathbf{v}}_i,
\end{equation}
where $\Delta\mathbf{R}_{ij}$ brings the velocity at $j$ into the frame at $i$.
We match this prediction to the integrated increment $\Delta\mathbf{v}_{ij}$ with
\begin{equation}
\label{eq:consistency_loss}
\mathcal{L}^{\mathrm{int}}=\sum_{(i,j)} \left\|\Delta\hat{\mathbf{v}}^{\mathrm{net}}_{ij}-\Delta\mathbf{v}_{ij}\right\|_2^2.
\end{equation}

The sum runs over sampled training windows.
\subsection{Velocity Regularization}
\label{sec:stationary}
At rest, residual noise and bias produce small nonzero integration targets that the network would otherwise reproduce as spurious motion.
Static frames supply a label-free zero-velocity constraint, and detecting them from the raw IMU is the most straightforward signal available on every platform, so we adopt it as the regularizer.
We detect static frames from low variation in accelerometer and gyroscope magnitudes, a small mean angular rate, and an acceleration magnitude close to gravity.
On these frames, $\mathcal{L}^{\mathrm{reg}}$ penalizes the mean squared predicted speed and is added to the integration loss with a weighting coefficient.
The same frames provide the teacher's constant bias estimates by averaging deviations from zero angular velocity and the expected gravity reading.

\subsection{Pose Graph Distillation}
\label{sec:m2}

\addtocounter{table}{-2}
\begin{table}[t]
\centering
\captionsetup{font=small}
\vspace{6pt}
\caption{\textbf{Evaluation datasets.}
Six datasets on six distinct embodiments, spanning different IMU grades and motion patterns.
}
\label{tab:dataset_statistics}
\resizebox{0.8\columnwidth}{!}{%
\begin{tabular}{l c c c}
\toprule
Dataset & Time & Length & Max Speed \\
\midrule
\logo{figs/logos/TLIO_logo.pdf}
TLIO~\citep{liu2020tlio}                & 31.6\,{\small h}    & 42.3\,{\small km}   & 6.0\,{\small m/s} \\
\logo{figs/logos/EuRoC_logo.pdf}\!\!
EuRoC~\citep{burri2016euroc}            & 22.5\,{\small min}  & 892\,{\small m}     & 2.9\,{\small m/s} \\
\logo{figs/logos/BlackBird_logo.pdf}
Blackbird~\citep{antonini2018blackbird} & 5.0\,{\small h}     & 28.4\,{\small km}   & 8.7\,{\small m/s} \\
\logo{figs/logos/MIPO_logo.pdf}
MIPO~\citep{yang2023mipo}               & 58.6\,{\small min}  & 1.4\,{\small km}    & 6.6\,{\small m/s} \\
\logo{figs/logos/ANYmal_logo.pdf}
GrandTour~\citep{frey_tuna2026grandtour} & 5.65\,{\small h}   & 10.3\,{\small km}   & 1.5\,{\small m/s} \\
\logo{figs/logos/Humanoid_logo.pdf}
Humanoid                                & 1.31\,{\small h}  & 1.9\,{\small km}     & 3.9\,{\small m/s} \\
\bottomrule
\end{tabular}}
\vspace{0pt}
\end{table}

\addtocounter{table}{1}
\begin{table*}[!t]
\centering
\begin{minipage}[t]{0.56\textwidth}
\vspace{0pt}
\centering
\captionsetup{type=table,font=small}
\caption{
\textbf{Self-Supervision with Proprioceptive Input.}
We report AVE (m/s) and ATE (m) on three legged platforms, with and without joint states added to the base IMU.
}
\label{tab:proprio_eval}
{\footnotesize
\resizebox{\linewidth}{!}{%
\begin{tabular}{l cc cc cc}
\toprule
 & \multicolumn{2}{c}{\logo{figs/logos/MIPO_logo.pdf}MIPO}
 & \multicolumn{2}{c}{\logo{figs/logos/ANYmal_logo.pdf}GrandTour}
 & \multicolumn{2}{c}{\logo{figs/logos/Humanoid_logo.pdf}Humanoid} \\
\cmidrule(lr){2-3} \cmidrule(lr){4-5} \cmidrule(lr){6-7}
Method & AVE~$\downarrow$ & ATE~$\downarrow$ & AVE~$\downarrow$ & ATE~$\downarrow$ & AVE~$\downarrow$ & ATE~$\downarrow$ \\
\midrule
\multicolumn{7}{l}{\textbf{Model-Based Method}} \\
\hspace{3mm} MIPO (IEKF)~\citep{yang2023mipo}
  & 0.199  & 0.632
  & --     & --
  & --     & -- \\
\midrule
\multicolumn{7}{l}{\textbf{Supervised Learning}} \\
\hspace{3mm} \airiob{} - \textit{baseline}
  & 0.121  & 0.310
  & 0.070  & 3.380
  & 0.096  & 1.400 \\
\hspace{3mm} \airiob{} + \textit{Proprio}
  & 0.116  & 0.190
  & 0.073  & 3.620
  & 0.095  & 2.350 \\
\midrule
\multicolumn{7}{l}{\textbf{Self-Supervised From Scratch (ours)}} \\
\rowcolor{oursbg} \hspace{3mm} Ours
  & \dcell{0.182}{+0.061}{dgreen} & \dcell{1.630}{+1.320}{dgrey}
  & \dcell{0.096}{+0.026}{dgreen} & \dcell{4.370}{+0.990}{dgreen}
  & \dcell{0.167}{+0.071}{dgreen} & \dcell{4.360}{+2.960}{dgrey} \\

\rowcolor{oursbg} \hspace{3mm} Ours + \textit{Proprio}
  & \dcell{0.146}{+0.030}{dgreen} & \dcell{0.820}{+0.630}{dgreen}
  & \dcell{0.091}{+0.018}{dgreen} & \dcell{3.660}{+0.040}{dgreen}
  & \dcell{0.161}{+0.066}{dgreen} & \dcell{3.200}{+0.850}{dgreen} \\

\bottomrule
\end{tabular}%
}}
\par\vspace{3pt}
{\footnotesize\parbox{\linewidth}{\raggedright
``--'' marks a platform that lacks the sensors a row requires.
\textit{+Proprio} adds the joint states to the single base IMU and, on MIPO, also the four calf-mounted IMUs.
Parenthesized numbers give the difference to the supervised row with the same input. \textcolor{dgreen}{Green} marks a difference below $+0.1$~m/s in AVE or $+1$~m in ATE; \textcolor{dgrey}{gray} a larger one.}}
\end{minipage}%
\hfill
\begin{minipage}[t]{0.43\textwidth}
\vspace{0pt}
\centering
\captionsetup{type=figure,font=small}
\captionof{figure}{\textbf{Top: Proprioceptive Input} for Go1 (MIPO), ANYmal (GrandTour), and G1 (Humanoid). \textbf{Bottom: GrandTour Trajectories} for the \texttt{EIG-1} and \texttt{SNOW-2} sequences.}
\label{fig:proprio_embodiments}
\includegraphics[width=\linewidth]{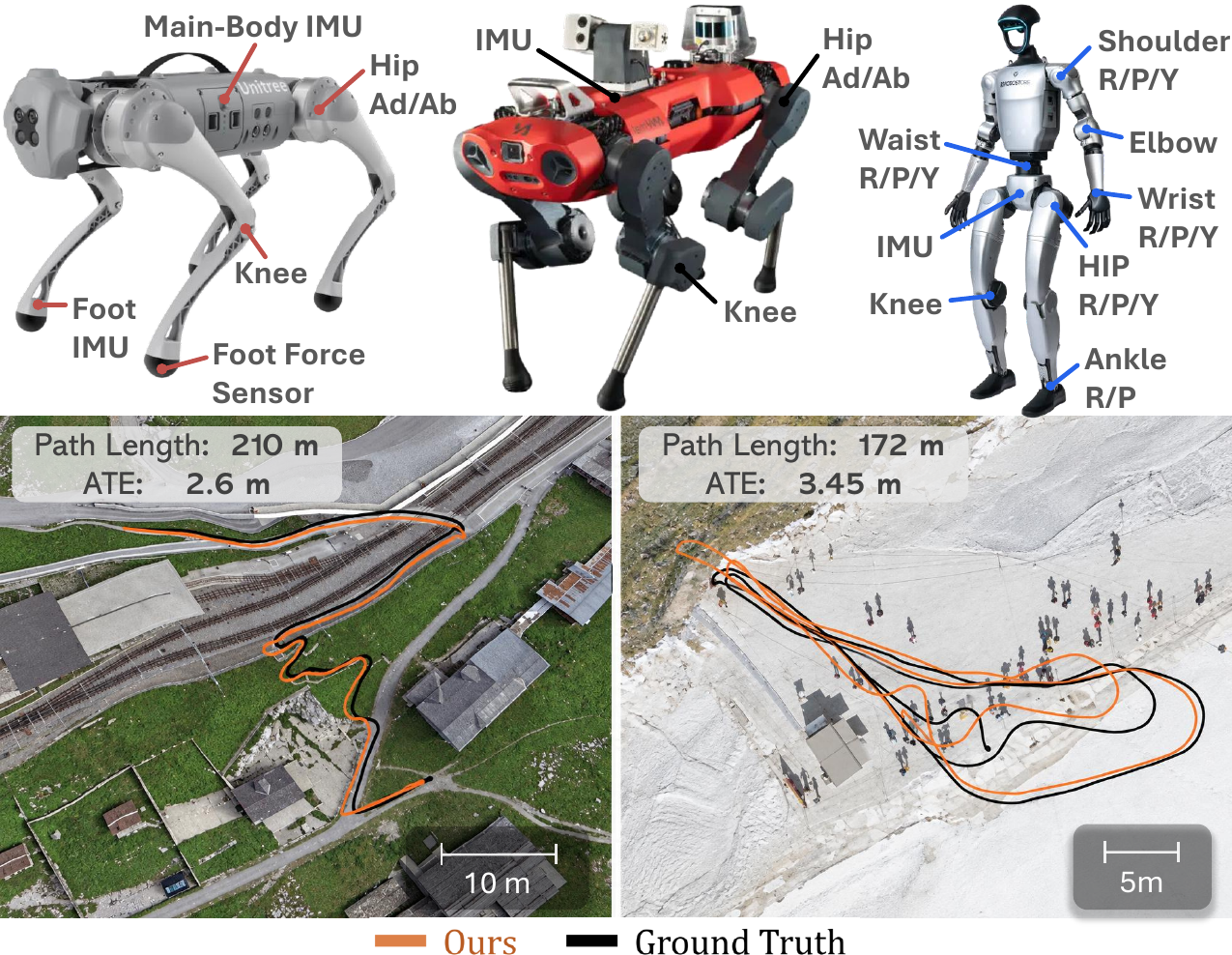}
\end{minipage}

\end{table*}

A single-window target carries one window of teacher noise and no consistency across windows.
Chaining many windows in a pose graph averages that noise and yields a velocity posterior with a longer effective horizon than any single window, a sharper target from the same teacher.
PGD starts from an integration-self-supervised network because the graph also consumes network velocity predictions.
Because the graph adds no independent measurement, repeating the round copies the student's own drift back into the target (\tref{tab:stage_ablation}), so we optimize each training graph once and fine-tune the network on the resulting targets, unlike the repeated alternation of imperative learning~\citep{wang2024imperative}.

The graph jointly optimizes orientation and body-frame velocity at its keyframes, with the first keyframe held fixed to remove the gauge freedom of the graph.
Integration factors enforce consistency with the teacher's relative rotations and velocity changes, while network factors keep velocities close to the student's predictions.
We weight integration factors by their propagated uncertainty and use a fixed isotropic weight for network factors, since self-supervision does not train the predicted covariance.
Position is omitted because it provides no additional velocity constraint.
We solve this weighted least-squares problem with Gauss--Newton, using the sparse solver of \aref{app:efficient_impl}.
We forward-integrate corrected IMU between the optimized keyframes to obtain dense velocity targets $\{{}^\mathcal{B}\mathbf{v}^{\mathrm{opt}}_k\}$ at every sample.
PGD minimizes
\begin{equation}
\mathcal{L}^{\mathrm{PGD}}
=
\sum_{k}\|{}^\mathcal{B}\mathbf{v}^{\mathrm{opt}}_k-{}^\mathcal{B}\hat{\mathbf{v}}_k\|_2^2,
\label{eq:pgd_loss}
\end{equation}
so the network learns the longer-horizon velocity that the graph recovers from the same integration teacher.

\subsection{Proprioceptive Input}
\label{sec:proprio_input}

The proprioceptive variant, reported as \airiob{} + \textit{Proprio} when trained with labels and Ours + \textit{Proprio} when self-supervised, extends the same network to read the articulated sensing available on legged platforms.
The student also encodes joint positions, velocities, and torques, together with auxiliary IMUs when available.
On MIPO~\citep{yang2023mipo}, these include four calf-mounted IMUs in addition to the base IMU.
The additional inputs require no annotation and expose embodiment context such as gait phase and contact.
Both variants use the same teacher, losses, and training procedure, so any gain from proprioception reflects student-side information rather than a stronger teacher.

\section{Experiments}
\label{sec:experiments}

We first evaluate self-supervised training from scratch across all six datasets, then isolate the effect of proprioceptive inputs on the three articulated platforms.

\textbf{Integration teacher.}
We use the standard IMU preintegrator of \sref{sec:m1}, with constant biases estimated from static frames; the already calibrated TLIO and MIPO measurements require no additional correction.
Bias correction affects only the training targets, while the student consumes raw IMU measurements.
The integration loss sums velocity-change residuals over a geometric series of gaps, from a single network step up to a few seconds.
The short gaps keep the teacher nearly exact while the long ones make the target informative about motion rather than noise; the largest gap is chosen per dataset so that the teacher's integration error over it stays below the target velocity error (\fref{fig:window_ablation}).
The static-frame detector of \sref{sec:stationary} uses one set of thresholds for the regularization mask on every dataset: over a centered window of $1$~s, the rolling standard deviations of the specific-force and angular-rate magnitudes must fall below $0.20\,\mathrm{m/s^2}$ and $0.05\,\mathrm{rad/s}$, the mean angular rate below $0.10\,\mathrm{rad/s}$, and the mean specific force within $0.40\,\mathrm{m/s^2}$ of gravity; only the regularization weight is tuned per dataset.
The bias estimate reuses this mask, so the teacher needs no calibration step beyond the static frames already detected for regularization.
In \tref{tab:setup2} and \tref{tab:proprio_eval}, for ease of implementation and to keep the protocol identical to AirIO, the teacher initializes the gravity direction from each dataset's reference orientation at the window start and propagates rotation with the bias-corrected gyroscope within the window.
This is an experimental choice, not a requirement of the method: \tref{tab:tilt_ablation} (\sref{app:limitations}) replaces the reference attitude with a gyroscope-integrated attitude on TLIO and the onboard AHRS on MIPO, and self-supervision still works.
PGD also needs an initial state for each graph, so we set the rotation and velocity of the first frame of every sequence from the reference trajectory.
No velocity or position labels serve as training targets, and the teacher requires no separately trained sensor model.

\textbf{Datasets.}
We evaluate on six datasets covering distinct embodiments, all summarized in \tref{tab:dataset_statistics}: {TLIO}~\citep{liu2020tlio} (pedestrians), {EuRoC}~\citep{burri2016euroc} (micro aerial vehicles), {Blackbird}~\citep{antonini2018blackbird} (racing drones), {MIPO}~\citep{yang2023mipo} (a \texttt{\small Unitree Go1} quadruped indoors), {GrandTour}~\citep{frey_tuna2026grandtour} (an \texttt{\small ANYmal} quadruped on in-the-wild outdoor terrain), and our own {Humanoid} set.
The set spans pedestrians, drones, quadrupeds, and a humanoid and stresses different regimes: Blackbird reaches $8.7$~m/s in aggressive flight, MIPO runs fast gaits up to $6.6$~m/s, and GrandTour covers $10.3$~km of outdoor terrain.
The two quadruped datasets are complementary: MIPO covers fast indoor gaits over short sequences with motion-capture ground truth, while GrandTour covers long outdoor missions with GNSS-inertial ground truth, so together they separate gait-induced IMU excitation from long-horizon drift.
The Humanoid data are collected on a \texttt{\small Unitree G1} using LiDAR-inertial odometry~\citep{zhao2021superodometryimucentriclidarvisualinertial} and learning-based stereo visual odometry~\citep{qiu2025macvo} as the source of ground truth.

\textbf{Metrics.}
\label{sec:metrics}
We report the average velocity error (AVE) as the primary metric, since our method directly trains body-frame velocity, and the absolute trajectory error (ATE) after integration or filtering as a secondary metric.
AVE is the mean Euclidean error in body-frame velocity.
ATE is the root-mean-square position error after a single rigid alignment of the estimated trajectory to ground truth, without scale adjustment.
Following the network-only protocol of AirIO~\citep{qiu2025airio}, ATE for every learned method integrates the predicted body-frame velocity with the ground-truth orientation, so supervised and self-supervised rows are compared under identical conditions.

\subsection{Self-Supervised Learning from Scratch}
\label{sec:tlio_result}

Self-supervision comes close to supervised learning on every embodiment.
With the reference-attitude teacher of \tref{tab:setup2} and PGD, the AVE gap to \airiob{} stays below $0.07$~m/s on every dataset.
\tref{tab:setup2} marks in \textcolor{dgreen}{green} every self-supervised cell within $0.1$~m/s of AVE or $1$~m of ATE of the supervised baseline.
The ATE gap is largest on EuRoC and Blackbird, where a constant velocity offset invisible to the velocity-change objective integrates into position error (\fref{fig:noise_bias_surface}).

The self-supervised model also outperforms some supervised methods.
Ours (w/ PGD) reaches lower AVE than both the supervised TLIO and RoNIN networks on five of the six datasets, all except MIPO.

Integration self-supervision alone already delivers most of this accuracy.
PGD then lowers AVE by $4$--$11\%$ on five of the six datasets by extending supervision from a single window to the longer horizon of the graph posterior (\sref{app:limitations}); on GrandTour the gain is marginal ($0.096$ to $0.095$~m/s).
ATE still rises on EuRoC ($9.03$ to $10.23$~m), where early velocity error that integration carries forward outweighs the AVE gain, and slightly on MIPO ($1.63$ to $1.64$~m).

\subsection{Self-Supervision with Proprioceptive Input}
\label{sec:setup3}

Proprioceptive input strengthens the student even though the teacher and loss are unchanged.
We test this on three articulated platforms that log proprioception: MIPO~\citep{yang2023mipo} on a Unitree Go1, GrandTour~\citep{frey_tuna2026grandtour} on ANYmal, and our humanoid set.
We train the student with the base IMU alone and with the available proprioception (\textit{proprio}), the joint states plus, on MIPO, the four calf-mounted IMUs (\fref{fig:proprio_embodiments}).
Proprioception lowers both AVE and ATE on all three platforms (\tref{tab:proprio_eval}), and on MIPO it puts the student within 0.03~m/s of its supervised twin.

On MIPO the natural reference is the model-based Invariant Extended Kalman Filter (IEKF) of the MIPO paper, which is tuned for this robot and fuses every proprioceptive stream through a hand-specified leg-kinematics model.
The self-supervised student exceeds it in velocity accuracy with neither a kinematic model nor a label: the base-IMU student already reaches a lower AVE ($0.182$ vs. $0.199$~m/s), and the proprioceptive student widens the margin to $27\%$ ($0.146$~m/s).
Only the filter's ATE stays lower; the filter estimates its own orientation, unlike the learned rows (\sref{sec:metrics}).

GrandTour shows that the recipe holds up outdoors over long distances.
\fref{fig:proprio_embodiments} shows two of these missions, \texttt{EIG-1} (210~m) and \texttt{SNOW-2} (172~m): trained only on the robot's own recordings, the proprioceptive student's integrated path follows the GNSS-inertial ground truth over the full route in both, including on snow.

\section{Analysis of the Learned Velocity Model}
\label{sec:ablation}

\definecolor{worse}{HTML}{B30326}
\definecolor{better}{HTML}{3A4CC0}

\begin{figure}[t]
\centering
\includegraphics[width=0.85\linewidth]{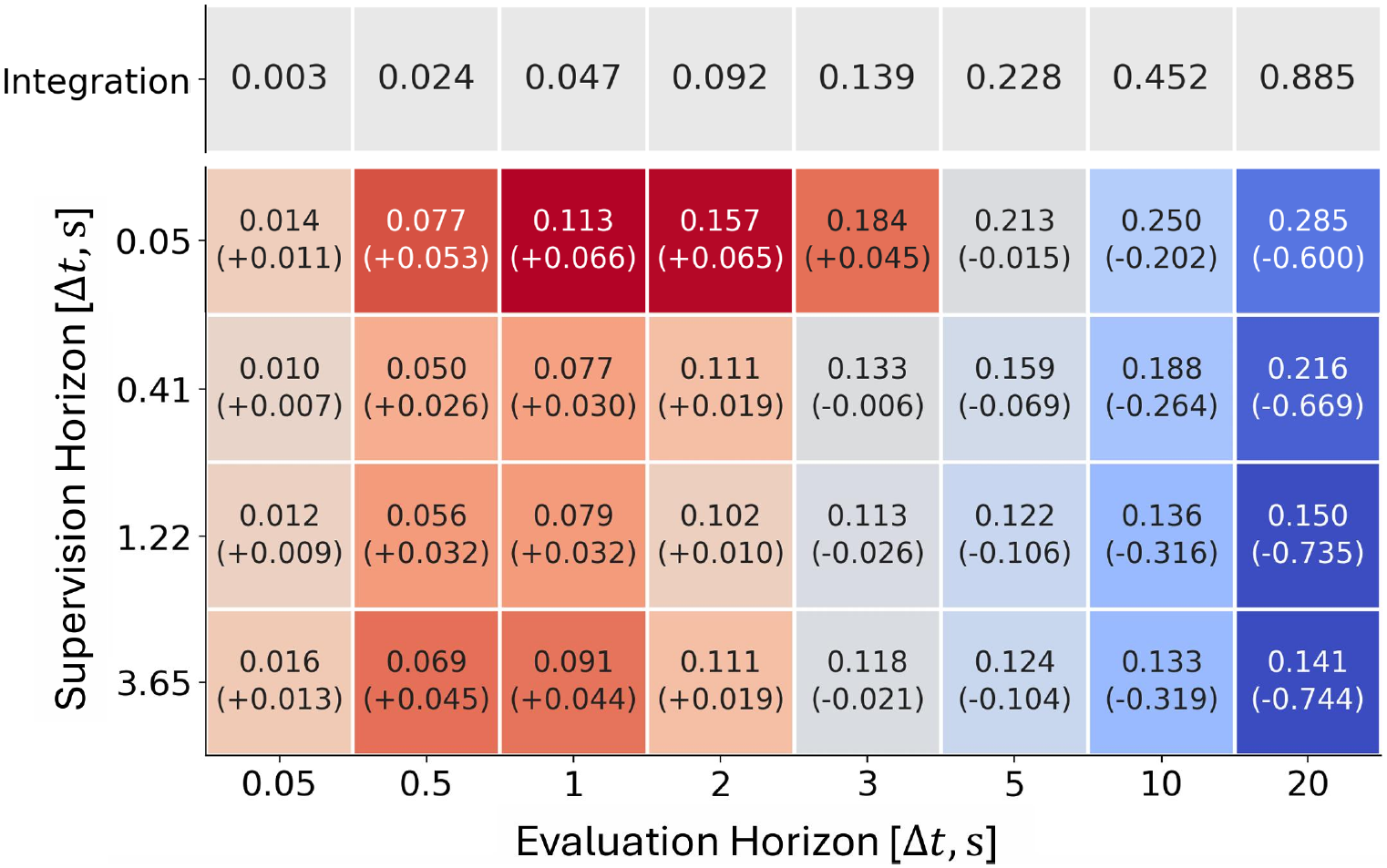}
\caption{Supervision window ablation on TLIO.
Velocity-change error $\|\Delta\hat{\mathbf{v}}^{\mathrm{net}}_{ij}-\Delta\mathbf{v}^\star_{ij}\|_2$ [m/s] over evaluation and training horizons, where $\Delta\mathbf{v}^\star_{ij}$ is the ground-truth body-frame velocity change;
{\color{worse}red} cells underperform and {\color{better}blue} cells outperform the integrator.}
\label{fig:window_ablation}
\vspace{-5pt}
\end{figure}

The integration teacher supplies every velocity-change target used to train the student.
Because each target is obtained by integrating the IMU over one window, a natural expectation is that the student learns to reproduce the teacher's integration behavior.
However, the trained student achieves lower velocity-change error than the teacher over long evaluation horizons (\fref{fig:window_ablation}).
This raises the question: \textbf{what has the student learned beyond the teacher's per-window integration?}

We interpret this behavior as evidence that the student learns an embodiment-specific motion model, using regularities in the platform's IMU history to predict body-frame velocity.
The teacher integrates each window independently, while the student fits a shared velocity predictor across many windows and can exploit recurring motion patterns.
We examine this interpretation through three analyses: the student's error across evaluation horizons, its transfer across datasets, and its response to additional unlabeled training data.
The horizon and data-volume studies use TLIO because it has the largest training set ($25$~h) and supports a wide range of supervision horizons.

\textbf{Does the student reproduce the teacher's integration behavior?}
The student is trained to match the teacher's integrated velocity changes, so a natural expectation is that it learns the teacher's integration function.
If so, its error should grow with integration time in a similar way to the teacher's.
We test this expectation by comparing their velocity-change errors as the evaluation horizon extends beyond the supervision horizon.

We train the network with supervision horizons from $0.05$~s to $3.65$~s and measure both student and teacher velocity-change error against ground truth across evaluation horizons (\fref{fig:window_ablation}).
The teacher is more accurate over short horizons, but its error grows with integration time.
The student is less accurate locally, yet its error grows more slowly.
At $20$~s, the model trained with the longest supervision window reaches approximately $0.14$~m/s error, compared with the teacher's $0.89$~m/s.
This roughly sixfold reduction shows that the student can produce more accurate long-horizon velocity changes than the integration procedure that supplied its training targets.

\begin{figure}[t]
\centering
\includegraphics[width=0.86\linewidth]{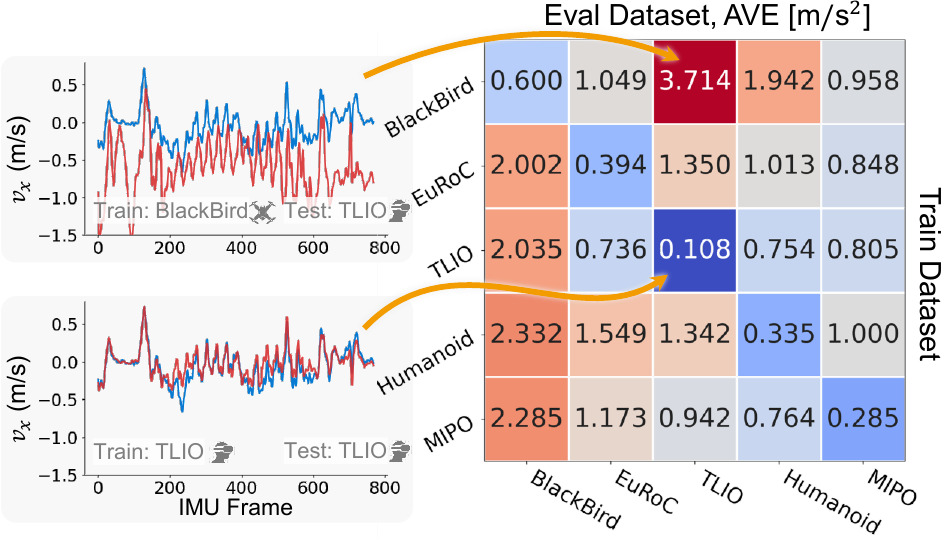}
\captionsetup{font=small}
\caption{\textbf{Cross-dataset transfer over five datasets.}
Each cell reports AVE for a model trained on one dataset and evaluated on another.
Models trained on the target dataset outperform transferred models, showing the importance of training on the target distribution.
The traces show a representative TLIO--Blackbird mismatch.}
\label{fig:cross_embodiment}
\vspace{-5pt}
\end{figure}

\begin{figure*}[t]
\centering
\begin{minipage}[t]{0.38\textwidth}
\vspace{0pt}
\centering
\includegraphics[width=0.79\linewidth]{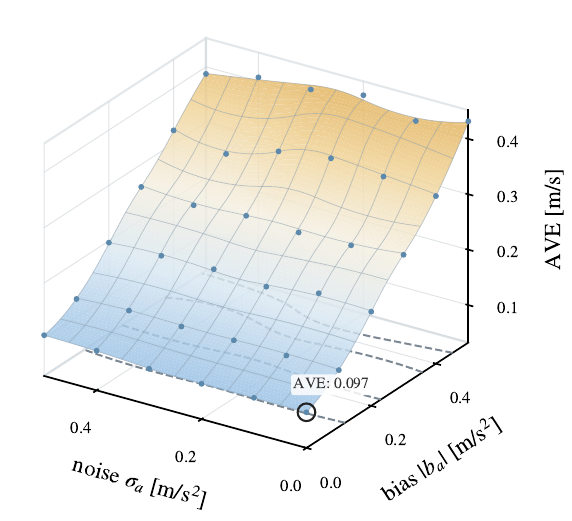}
\captionsetup{font=small}
\caption{\textbf{Noise and bias ablation on TLIO.}
Held-out AVE after self-supervised training with zero-mean Gaussian noise $\sigma_a$ and a constant bias $\|\mathbf{b}_a\|_2$ injected into the training IMU (gyroscope perturbed at one tenth of each value in $\mathrm{rad/s}$).}
\label{fig:noise_bias_surface}
\end{minipage}%
\hfill
\begin{minipage}[t]{0.6\textwidth}
\vspace{0pt}
\centering
\includegraphics[width=\linewidth]{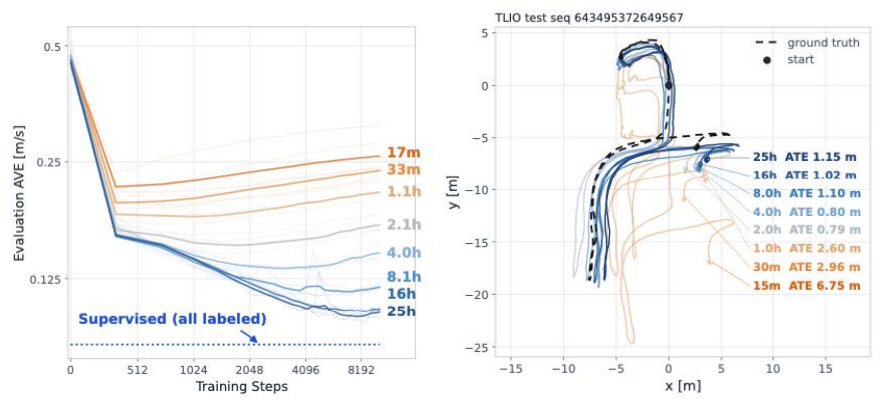}
\captionsetup{font=small}
\caption{\textbf{Scaling with unlabeled TLIO data.}
\textbf{Left:} held-out AVE over training steps for nested subsets of the training split, from a quarter hour to the full $25$~h, against the supervised \airiob{} baseline (dotted).
\textbf{Right:} integrated trajectory on a held-out TLIO sequence for each subset with its ATE; the dashed line is the ground truth.
More unlabeled data lowers both toward the supervised baseline.}
\label{fig:data_scaling}
\vspace{-5pt}
\end{minipage}
\end{figure*}

\textbf{Does the learned model transfer across datasets?}
IMU integration follows the same physical equations across platforms.
If the student learns this integration function, we would expect it to transfer across datasets.
A learned motion model, however, may depend on the motion patterns seen during training.
We examine this dependence by evaluating each trained model on other datasets without retraining (\fref{fig:cross_embodiment}).

Every transferred model has at least $1.9\times$, and up to $34\times$, the AVE of the model trained on the evaluation dataset.
The degradation also occurs between the two aerial datasets: the Blackbird-trained model more than doubles the EuRoC-trained model's error on EuRoC.
The TLIO--Blackbird velocity traces illustrate this mismatch at the sequence level.
These results support training on the target platform's data, but do not separate embodiment effects from differences in sensors, calibration, or motion distributions.

\textbf{Does accuracy scale with unlabeled data?}
We train the same pipeline on nested subsets of the $25$~h TLIO training split, from a quarter hour up to the full split with three seeds each, and evaluate every checkpoint on a fixed held-out split (\fref{fig:data_scaling}).
Held-out AVE improves monotonically with more unlabeled data and approaches the supervised \airiob{} baseline at the full scale.

\section{Limitations}
\label{app:limitations}

Self-supervision has two requirements.
First, the training data must contain enough motion excitation, so the student observes diverse motions and learns their patterns.
Second, the IMU bias must be corrected, since a large uncorrected bias keeps the model from converging.

\textbf{Sensitivity to IMU quality and calibration bias.}
The student is robust to zero-mean noise but sensitive to bias.
We inject zero-mean Gaussian noise, a per-trajectory constant bias, or both into the training IMU measurements, train integration self-supervision from scratch on TLIO at each level, and report held-out AVE (\fref{fig:noise_bias_surface}).
Noise with a standard deviation of up to $0.5\,\mathrm{m/s^2}$, well beyond typical sensor noise, raises AVE by at most $9\%$, because zero-mean error largely averages out within and across windows.
A constant bias of $0.1\,\mathrm{m/s^2}$ already raises AVE by $39\%$ and a bias of $0.5\,\mathrm{m/s^2}$ more than quadruples it, because a systematic error accumulates within every window and never averages out.
Calibration therefore matters far more than sensor noise, which is why the teacher removes the bias estimated from static frames (\sref{sec:m1}).

\textbf{Sensitivity to tilt error.}
The teacher removes gravity with the attitude at each window start, so a tilt error leaks gravity into every target.
\tref{tab:tilt_ablation} replaces the reference attitude in the teacher with one obtained from the IMU alone on two datasets chosen for this ablation: TLIO, because it has enough data and a clean, well-calibrated IMU, and MIPO, because it is the only dataset that provides an onboard AHRS.
On TLIO we integrate the raw gyroscope forward from the attitude at the sequence start, with no re-anchoring.
We clip every TLIO sequence to its first $2$~min and retrain the reference arm on the same clips so that the two columns differ only in the attitude source.
On MIPO the teacher uses the attitude from the onboard AHRS of the \texttt{\small Unitree Go1}, which sits at a mean tilt of $0.9^\circ$ from the motion-capture attitude.
Self-supervision still works from either source: the gyroscope attitude raises AVE by $22\%$ on TLIO and the AHRS attitude by $29\%$ on MIPO, the latter matching the $0.3$~m/s of gravity that a $0.9^\circ$ tilt leaks into a target window.
The loss tracks the tilt error of the attitude source, so a better source would narrow the gap further: finer gyroscope bias calibration or a gravity-aided attitude filter would hold the tilt below what raw integration or a consumer AHRS achieves, but both are outside the scope of this paper.
\begin{table}[t]
    \centering
    \captionsetup{font=small}
    \caption{\textbf{Attitude source in the teacher.}
    Test AVE (m/s) / ATE (m) of Ours with the reference tilt and with a tilt estimated from the IMU alone: the gyroscope integrated from the sequence start on TLIO, the onboard AHRS on MIPO. Parentheses: AVE change relative to the reference column.}
    \label{tab:tilt_ablation}
    \footnotesize
    \setlength{\tabcolsep}{4pt}
    \begin{tabular}{l c c c}
    \toprule
    Dataset (source) & Reference tilt & Estimated tilt & Tilt error \\
    \midrule
    \logo{figs/logos/TLIO_logo.pdf}TLIO {\scriptsize(gyro)} & 0.109 / 3.21 & 0.133 / 3.81 (+22\%) & $<1^\circ$ \\
    \logo{figs/logos/MIPO_logo.pdf}MIPO {\scriptsize(AHRS)} & 0.182 / 1.63 & 0.235 / 1.41 (+29\%) & $0.9^\circ$ \\
    \bottomrule
    \vspace{0pt}
    \end{tabular}
    \par\vspace{10pt}
    \captionsetup{font=small}
    \caption{\textbf{PGD ablation on GrandTour.}
    Test AVE (m/s) / ATE (m) of the checkpoint after the last label iteration.
    The from-scratch IMU-only run diverges after about 300 iterations.}
    \label{tab:stage_ablation}
    \vspace{4pt}
    \footnotesize
    \begin{tabular}{l c c}
    \toprule
    Initialization / label iterations & IMU only & + \textit{Proprio} \\
    \midrule
    \multicolumn{3}{l}{\textit{From integration self-supervision}} \\
    \hspace{3mm}(i) 1 iteration (ours) & 0.095 / 4.34 & 0.090 / 3.67 \\
    \hspace{3mm}(ii) 10 iterations     & 0.149 / 8.99 & 0.128 / 6.99 \\
    \hspace{3mm}(iii) 20 iterations    & 0.168 / 10.68 & 0.144 / 10.64 \\
    \hspace{3mm}(iv) 50 iterations     & 0.192 / 13.13 & 0.184 / 22.52 \\
    \hspace{3mm}(v) 80 iterations      & 0.202 / 15.19 & 0.249 / 37.47 \\
    \midrule
    \multicolumn{3}{l}{\textit{From scratch}} \\
    \hspace{3mm}(vi) 150 iterations    & 0.216 / 15.53 & 0.127 / 8.50 \\
    \hspace{3mm}(vii) 600 iterations   & 2.36 / 532    & 0.345 / 23.65 \\
    \bottomrule
    \end{tabular}
    \vspace{0pt}
\end{table}

\textbf{Does PGD need the initialization?}
A self-supervision target carries at most one window of teacher error, which averages out across windows, whereas the pose graph chains integration factors across its whole span, so attitude error and the gravity it leaks into velocity grow with distance from the anchor and only the network's velocity factors resist the drift.
Because the graph adds no independent measurement, repeating PGD reinforces this drift rather than correcting it, so we run one round.
\tref{tab:stage_ablation} tests both choices on GrandTour by re-solving the label from the current network every 20 optimizer steps.
From the self-supervised network every added iteration raises AVE, from 0.095 to 0.202~m/s over 80 iterations, because each re-solve copies the student's drift into its target.
From scratch, student and label improve for a few thousand steps and then drift the same way, ending $1.4$--$2.3\times$ above row (i) after 150 iterations ($3{,}000$ steps) and, after 600 iterations ($12{,}000$ steps), diverging (IMU only) or degrading further (+\textit{Proprio}).

\section{Conclusion}
\label{sec:conclusion}

We have shown that a robot can learn its own inertial odometry model from unlabeled IMU data alone, and we draw two lessons from this result.
The first is practical: learned inertial odometry no longer needs labeled trajectories for every new embodiment, only a calibrated IMU and the platform's own unlabeled recordings.
The second is conceptual: a myopic teacher can train a student that is more accurate than itself.
The integration teacher is myopic and ignorant of the platform, and its increments are noisy.
Trained on many such increments, the student averages the noise away and keeps what repeats, and so learns an embodiment-specific motion model.
This averaging argument explains why the student can surpass the teacher, but not why matching velocity changes yields a motion model rather than an integrator; that evidence is empirical.
Which motion properties make an embodiment learnable from noisy increments, and how much unlabeled data this takes, remain open, and a theoretical account of the mechanism is our main direction for future work.

\bibliographystyle{bibliography/IEEEtran}
\bibliography{bibliography/references, bibliography/papers, bibliography/yuheng}

\clearpage
\appendices
\begin{figure*}[t]
    \centering
    \includegraphics[width=0.75\textwidth]{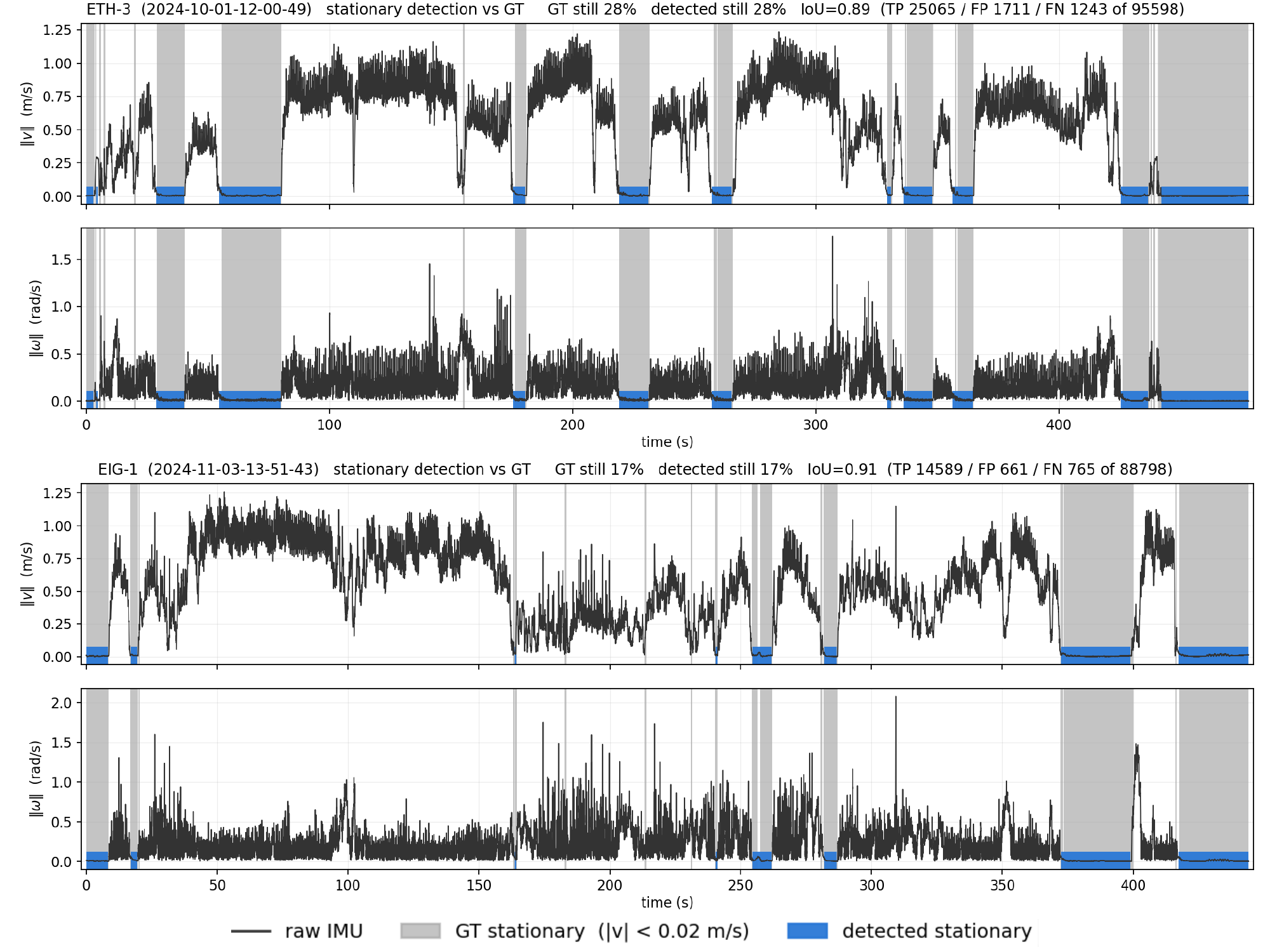}
    \caption{\textbf{Raw-IMU Stationary Window Detector} performance vs.\ GT stationary windows for two ANYmal GrandTour sequences, ETH-3 (top) and EIG-1 (bottom).}
    \label{fig:statomary_detect}
\end{figure*}

\section{Integration Covariance Propagation}
\label{app:airimu_covariance}

This section gives the covariance propagation matrices $\mathbf{A}_k$, $\mathbf{B}^{\mathrm{gyro}}_{k}$, and $\mathbf{B}^{\mathrm{acc}}_{k}$ behind the propagated uncertainty that weights the integration factors of pose graph distillation (\sref{sec:m2}).
The integration covariance is defined over the rotation and velocity errors $\delta\mathbf{z}_{ij}=[\delta\boldsymbol{\phi}_{ij}^{\top},\delta\mathbf{v}_{ij}^{\top}]^{\top}$:
\begin{equation}
\boldsymbol{\Sigma}^{\mathrm{int}}_{ij}
=
\begin{bmatrix}
\boldsymbol{\Sigma}^{\mathrm{int},\boldsymbol{\phi}}_{ij} & \boldsymbol{\Sigma}^{\mathrm{int},\boldsymbol{\phi}\mathbf{v}}_{ij}\\
\boldsymbol{\Sigma}^{\mathrm{int},\mathbf{v}\boldsymbol{\phi}}_{ij} & \boldsymbol{\Sigma}^{\mathrm{int},\mathbf{v}}_{ij}
\end{bmatrix}.
\label{eq:app_integration_cov_blocks}
\end{equation}
Following AirIMU, the state-transition matrix is
\begin{equation}
\mathbf{A}_{k}
=
\begin{bmatrix}
\Delta\mathbf{R}_{k,k+1}^{\top} & \mathbf{0}_{3\times3}\\
-\Delta\mathbf{R}_{ik}({}^{\mathcal{B}}\tilde{\mathbf{a}}_k)^{\wedge}\Delta t_k & \mathbf{I}_{3\times3}
\end{bmatrix}.
\label{eq:app_covariance_A}
\end{equation}
The gyroscope and accelerometer noise injection matrices are
\begin{equation}
\mathbf{B}^{\mathrm{gyro}}_{k}
=
\begin{bmatrix}
\mathbf{J}^{r}_{k}\Delta t_k\\
\mathbf{0}_{3\times3}
\end{bmatrix},
\qquad
\mathbf{B}^{\mathrm{acc}}_{k}
=
\begin{bmatrix}
\mathbf{0}_{3\times3}\\
\Delta\mathbf{R}_{ik}\Delta t_k
\end{bmatrix}.
\label{eq:app_covariance_B}
\end{equation}
Here, ${}^{\mathcal{B}}\tilde{\mathbf{a}}_k$ is the bias-corrected accelerometer sample, $(\cdot)^{\wedge}$ maps a vector to its skew-symmetric matrix and $\mathbf{J}^{r}_{k}$ is the right Jacobian of $SO(3)$ for the gyroscope increment at frame $k$.

\section{Training Details}
\label{app:training_details}
We report the training configuration for both components.
Integration self-supervision and pose graph distillation share the same motion network and optimizer family, and differ in the supervision signal and the parameters listed below.

\subsection{Integration Self-Supervision}
\label{app:training_stage1}
Integration self-supervision trains the network to match the integrator's velocity increments under the squared velocity-change loss in \eref{eq:consistency_loss}.
The main per-dataset settings are the batch size and the supervision horizon (the window length $|i-j|$ over which velocity increments are compared); datasets with similar dynamics share the same configuration.
The velocity regularization of \sref{sec:stationary} detects static frames (\fref{fig:statomary_detect}) with the window and thresholds given in \sref{sec:experiments}, shared by every dataset; only the weight $\lambda_{\mathrm{reg}}$ is tuned per dataset (\tref{tab:stage1_config}).

For the TLIO and GrandTour datasets, long supervision horizons of \texttt{5.12\,s} and \texttt{3.65\,s}, respectively, are adopted because the integration remains sufficiently accurate over long intervals. In contrast, the drone datasets (EuRoC and Blackbird) use shorter horizons, reflecting both the faster MAV motion and the increased sensitivity of integration errors during aggressive maneuvers. The MIPO dataset uses the shortest supervision horizon, \texttt{0.41\,s}, because of its noisy data and impact-rich quadruped locomotion. Similarly, the Humanoid dataset uses a shorter supervision horizon of \texttt{1.22\,s} to capture high-frequency bipedal motion while limiting the effect of integration inaccuracies.

\begin{table*}[t]
\centering
\captionsetup{font=small}
\caption{\textbf{Integration self-supervision training configuration.} Supervision horizon is the velocity-increment window length used during integration self-supervision.}
\label{tab:stage1_config}
\resizebox{\textwidth}{!}{%
\begin{tabular}{l c c c c c c c}
\toprule
Parameter & TLIO & EuRoC & Blackbird & MIPO & \multicolumn{2}{c}{GrandTour (quadruped)} & Humanoid \\
\cmidrule(lr){6-7}
 & (pedestrian) & (drone) & (drone) & (quadruped) & IMU-only & joints & \\
\midrule
Batch size              & \texttt{512}  & \texttt{64}   & \texttt{512}  & \texttt{256} & \texttt{256}  & \texttt{256} & \texttt{512} \\
Supervision horizon [s] & \texttt{5.12} & \texttt{1.22} & \texttt{2.43} & \texttt{0.41} & \texttt{3.65} & \texttt{3.65} & \texttt{1.22} \\
Learning rate           & \texttt{1e-3} & \texttt{1e-3} & \texttt{1e-3} & \texttt{5e-5} & \texttt{5e-5} & \texttt{5e-5} & \texttt{1e-3} \\
Epochs                  & \texttt{40}   & \texttt{150}  & \texttt{120}  & \texttt{240} & \texttt{120}  & \texttt{60}  & \texttt{128} \\
Regularization weight $\lambda_{\mathrm{reg}}$ & \texttt{0.25} & \texttt{0.1} & \texttt{0.1} & \texttt{0.01} & \texttt{1.0} & \texttt{0.1} & \texttt{1.0} \\
\bottomrule
\end{tabular}
}
\end{table*}


\subsection{Pose Graph Distillation}
\label{app:training_stage2}
Pose graph distillation seeds the pose graph with the integration self-supervised network, solves the graph, and distills the posterior velocity back into the network with the loss in \eref{eq:pgd_loss}.
We perform one outer PGD round.
Within this round, we build one graph per training sequence, optimize it once with Gauss--Newton (at most \texttt{10} iterations), cache the posterior velocity targets, and fine-tune the network on them.
We do not start another graph-distillation round from the refined checkpoint.
Each graph spans its full training sequence, with one keyframe per network output. The network is fine-tuned on windows of \texttt{1024}, which correspond to 5.12\,s for all datasets but Blackbird, where they correspond to 10.24\,s due to the lower sampling rate. 
For all datasets, we use a batch size of \texttt{512} and a learning rate of \texttt{5e-5}.

\subsection{Proprioceptive Input Fusion Architecture}
The motion network's encoders are 1D convolutions across the IMU and joint histories.
With proprioceptive input, the data from all four auxiliary IMU histories are flattened alongside the main-body IMU and fed to the IMU encoder, and the data from all joint and foot force sensor histories are flattened and fed to the optional joint encoder (\fref{fig:motion_network}).

\subsection{MIPO Data Preparation} \label{sec:mipo_data_prep}
This paper uses exclusively the indoor sequences of the MIPO dataset. For these, the ground truth pose is gathered by an indoor motion-capture system \cite{yang2023mipo}. For such systems, the poses may still have coordinate offsets and time synchronization offsets relative to the robot's world frame for each sequence. As such, Kalibr \cite{furgale2013kalibr} was used to calibrate the per-sequence extrinsics and time synchronization. In some cases, Kalibr was unable to converge to a solution. For these instances, the time offset is assumed to be zero. A small set of these sequences still had evident time synchronization issues, so they were excluded from the training, validation, and test sets.

\section{Efficient Implementation of Pose Graph Distillation}
\label{app:efficient_impl}

During the single outer PGD round, pose graph distillation solves one pose graph per
training sequence (\sref{app:training_stage2}), so its cost bounds how long a sequence
and how large a training set PGD can handle. Each graph carries a state
$\mathbf{x}=(\mathbf{x}_1,\dots,\mathbf{x}_N)\in\mathbb{R}^{6N}$ with
$\mathbf{x}_k=(\boldsymbol{\phi}_k,{}^{\mathcal{B}}\mathbf{v}_k)\in\mathbb{R}^{6}$,
and we take Gauss--Newton steps
\begin{equation}
\mathbf{H}\,\delta\mathbf{x}=-\mathbf{g},
\qquad
\mathbf{H}=\mathbf{J}^{\top}\mathbf{W}\mathbf{J},
\qquad
\mathbf{g}=\mathbf{J}^{\top}\mathbf{W}\mathbf{r},
\label{eq:app_gn_normal}
\end{equation}
batched over $B$ graphs.

\paragraph{Sparsity.}
The IMU preintegration factors are binary and couple only adjacent keyframes
$(k,k{+}1)$, while the velocity-observation and
velocity-prior factors are unary. Consequently,
$\mathbf{H}\in\mathbb{R}^{6N\times6N}$ is block-tridiagonal in $6\times6$ blocks,
\begin{equation}
\mathbf{H}_{k,l}=\mathbf{0}\ \ \text{for}\ \ |k-l|>1,
\qquad
\mathbf{H}_{k+1,k}=\mathbf{H}_{k,k+1}^{\top},
\label{eq:app_block_tridiag}
\end{equation}
so only $3N-2$ of the $N^2$ blocks are nonzero.

\paragraph{Cost.}
A dense solver discards this structure: it stores $\mathbf{H}$ and $\mathbf{W}$
densely and factorizes by Cholesky, requiring $\mathcal{O}(B\,N^{3})$ time and
$\mathcal{O}(B\,N^{2})$ memory. We instead assemble $\mathbf{H}$ as a batched
block-tridiagonal matrix in block sparse row (BSR) format with custom CUDA kernels on
NVIDIA Warp\footnote{\url{https://github.com/NVIDIA/warp}} and solve \eref{eq:app_gn_normal} with the
cuDSS sparse direct solver ($\mathbf{L}\mathbf{D}\mathbf{L}^{\top}$ factorization in double precision).
Because only the $\mathcal{O}(N)$ nonzero blocks are stored and factorized, each
Gauss--Newton step is $\mathcal{O}(B\,N)$ in memory and time and never
materializes the dense Hessian.

\paragraph{Results.}
Both backends converge in the same number of Gauss--Newton iterations to
identical estimates, with a maximum velocity disagreement of
$1.4\times10^{-5}$, so the sparse path is an exact acceleration rather than an
approximation. \tref{tab:efficient_impl} reports per-solve time at $B=128$. At
$N=128$ the sparse solver is $1.8\times$ faster. More consequentially, the
dense $\mathcal{O}(B\,N^{2})$ footprint exceeds the $32$~GB device for
$N\ge512$ (e.g., the dense $\mathbf{H}$ alone needs $36\,B\,N^{2}$ floats,
$4.8$~GB at $N{=}512$ before $\mathbf{W}$ and $\mathbf{J}$), whereas the sparse
solver scales linearly and solves $N=1024$ in under half a second.

\begin{table}[t]
\centering
\captionsetup{font=small}
\caption{\textbf{Pose-graph solve cost} at batch size $B=128$ on a single
NVIDIA RTX~5090 ($32$~GB), mean over $20$ solves. Both backends return the same
solution; \emph{OOM} marks configurations whose dense memory exceeds the
device.}
\label{tab:efficient_impl}
\resizebox{\columnwidth}{!}{%
\begin{tabular}{l c c c}
\toprule
Keyframes $N$ & Dense (Cholesky) & Sparse (BSR + cuDSS) & Speedup \\
\midrule
\texttt{128}  & \texttt{63.4\,ms} & \texttt{35.3\,ms}  & $1.8\times$ \\
\texttt{512}  & OOM               & \texttt{167.9\,ms} & n/a \\
\texttt{1024} & OOM               & \texttt{438.7\,ms} & n/a \\
\bottomrule
\end{tabular}%
}
\end{table}

\twocolumn[{%
  \section{Sample Trajectories}

  \begin{minipage}{\textwidth}
    \centering
    \includegraphics[width=0.88\linewidth,page=1]
      {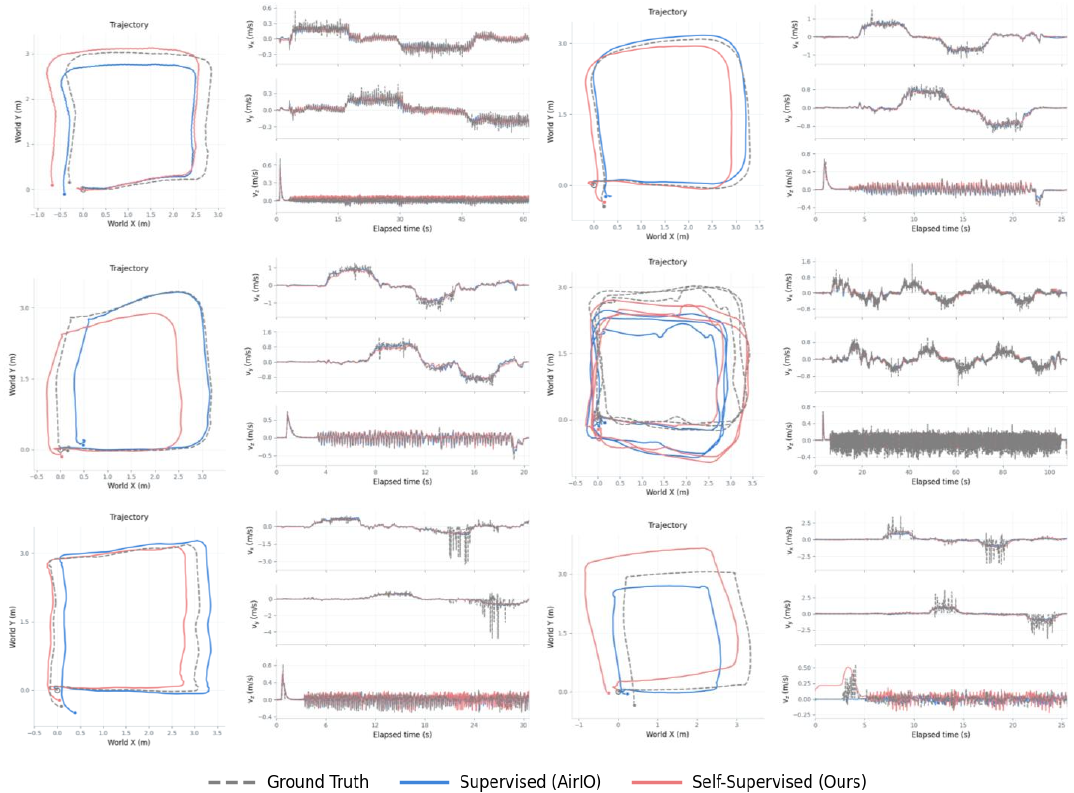}
    \captionof{figure}{Position and velocity on six sequences
      from the \textbf{MIPO} Test set.}
    \label{fig:mipo-results}

    \vspace{10pt}

    \includegraphics[width=0.88\linewidth,page=2]
      {figs/traj_gallery.pdf}
    \captionof{figure}{Position and velocity on six sequences
      from the \textbf{Blackbird} Test set.}
    \label{fig:blackbird-results}
  \end{minipage}
}]



\begin{figure*}[!t]
    \centering
    \includegraphics[width=0.86\linewidth, page=3]{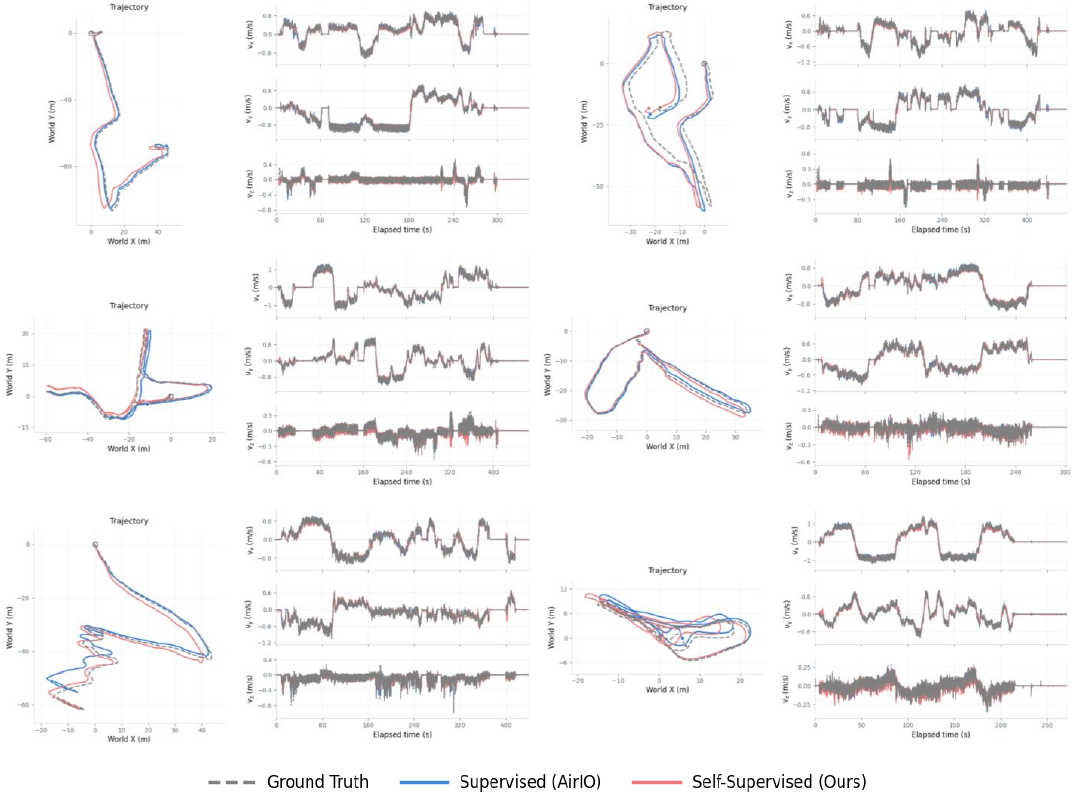}
    \vspace{-1pt}
    \caption{Position and velocity on six sequences from the \textbf{GrandTour} Test set.}
    \label{fig:grandtour-results}
\end{figure*}

\begin{figure*}[!t]
    \centering
    \includegraphics[width=0.86\linewidth, page=4]{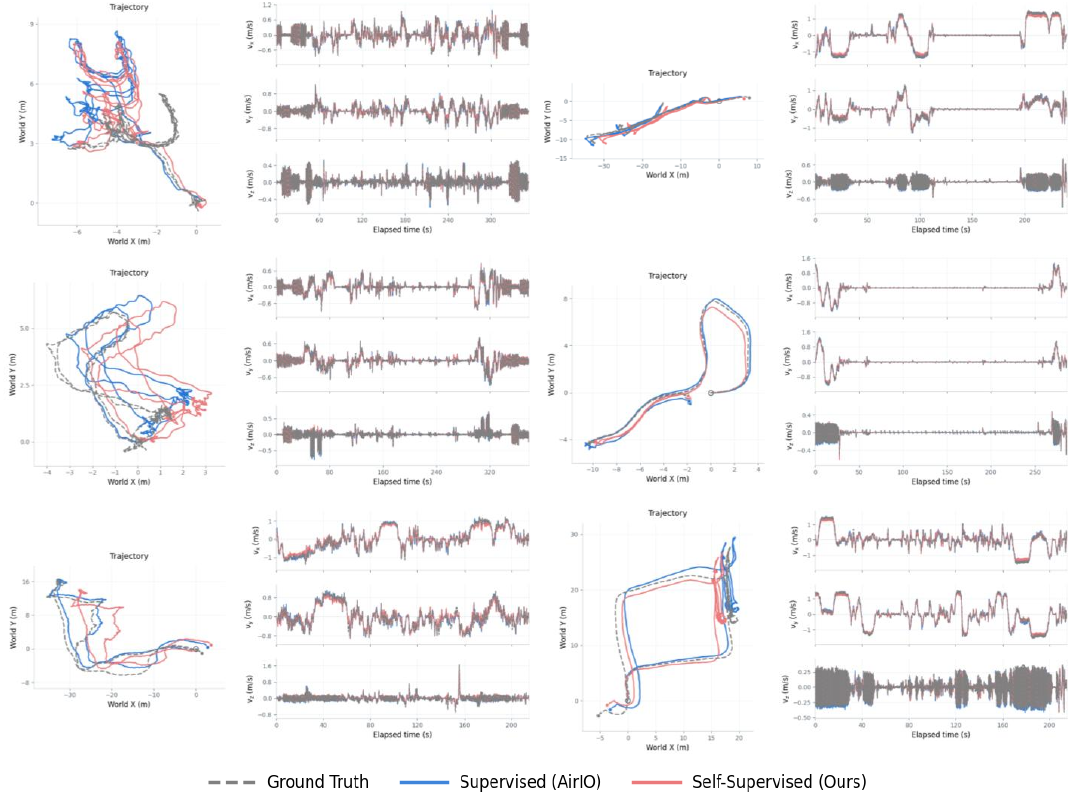}
    \vspace{-1pt}
    \caption{Position and velocity on six sequences from the \textbf{TLIO} Test set.}
    \label{fig:tlio-results}
\end{figure*}

\clearpage 

\makeatletter
\begingroup
\setlength{\@dblfptop}{0pt}

\begin{figure*}[!t]
    \centering
    \includegraphics[width=0.86\linewidth, page=5]{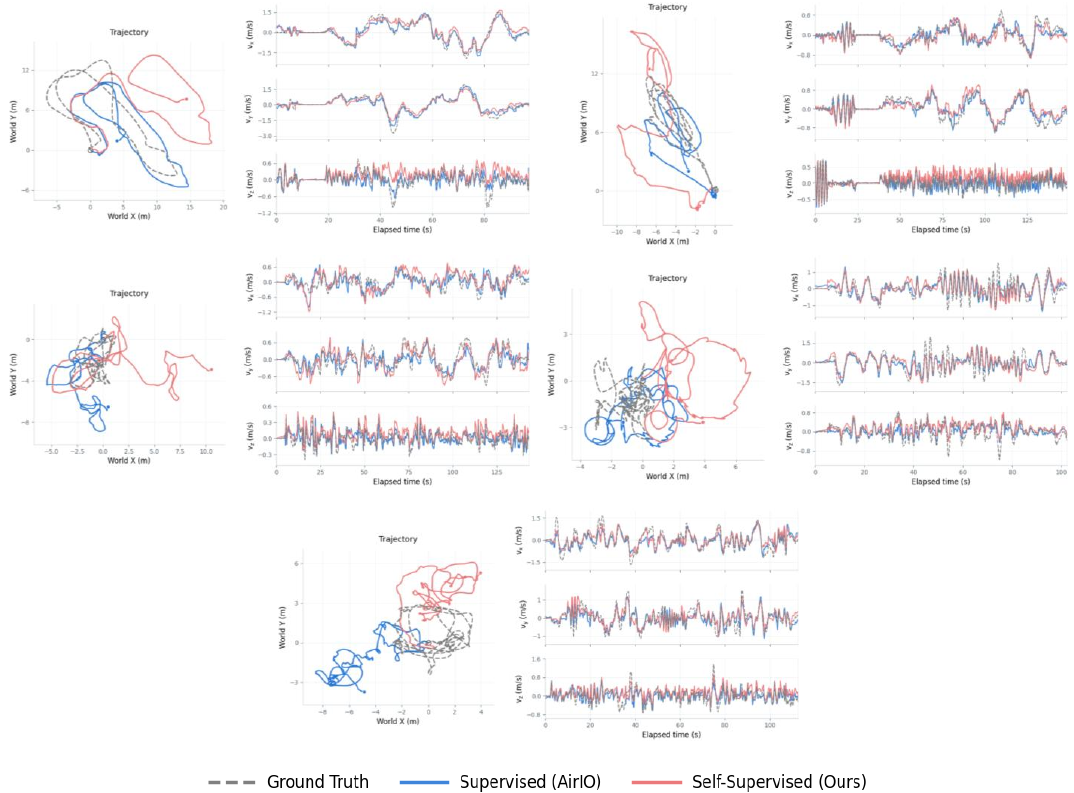}
    \vspace{-1pt}
    \caption{Position and velocity on six sequences from the \textbf{EuRoC} Test set.}
    \label{fig:euroc-results}
\end{figure*}

\clearpage 
\endgroup
\makeatother

\end{document}